\documentclass[final,1p,times,authoryear]{elsarticle}

\usepackage{amssymb}

\usepackage[utf8]{inputenc} 
\usepackage[sc]{mathpazo} 
\usepackage{amsmath}
\usepackage{amssymb}
\usepackage{mathtools} 
\usepackage{mathrsfs}  
\usepackage{xcolor}   
\usepackage[round]{natbib}
\usepackage[english]{babel} 
\usepackage[thmmarks,amsmath]{ntheorem} 
\usepackage{thmtools}

\usepackage{caption}
\usepackage{subcaption}
\usepackage{booktabs}

\usepackage[colorlinks,citecolor=blue,urlcolor=blue]{hyperref} 

\newcommand{\R}{{\mathbb{R}}}
\newcommand{\N}{{\mathbb{N}}}
\newcommand{\esp}{{\mathbb{E}}}

\newcommand{\prob}{{\mathbb{P}}}

\newcommand{\pkg}[1]{\texttt{#1}}

\newtheorem{proposition}{Proposition}

\newtheorem{definition}{Definition}

\journal{CSDA}

\begin{document}

\begin{frontmatter}



\title{Embedding and learning with signatures}


\author[label1]{Adeline Fermanian}
\address[label1]{Sorbonne Université, CNRS, Laboratoire de Probabilités, Statistique et Modélisation, 4 place Jussieu, 75005 Paris, France, adeline.fermanian@sorbonne-universite.fr}

\begin{abstract}
	Sequential and temporal data arise in many fields of research, such as quantitative finance, medicine, or computer vision. A novel approach for sequential learning, called the signature method and rooted in rough path theory, is considered. Its basic principle is to represent multidimensional paths by a graded feature set of their iterated integrals, called the signature. This approach relies critically on an embedding principle, which consists in representing discretely sampled data as paths, i.e., functions from $[0,1]$ to $\R^d$. After a survey of machine learning methodologies for signatures, the influence of embeddings on prediction accuracy is investigated with an in-depth study of three recent and challenging datasets. It is shown that a specific embedding, called lead-lag, is systematically the strongest performer across all datasets and algorithms considered. Moreover, an empirical study reveals that computing signatures over the whole path domain does not lead to a loss of local information. It is concluded that, with a good embedding, combining signatures with other simple algorithms achieves results competitive with state-of-the-art, domain-specific approaches. \\

\end{abstract}




\end{frontmatter}



%
    \section{Introduction}
    
    Sequential or temporal data are arising in many fields of research, due to an increase in storage capacity and to the rise of machine learning techniques. An illustration of this vitality is the recent relaunch of the Time Series Classification repository \citep{bagnalluea}, with more than a hundred new datasets. Sequential data are characterized by the fact that each sample consists of an ordered array of values. The order need not correspond to time, for example, text documents or DNA sequences have an intrinsic ordering, and are, therefore, considered as sequential. Besides, when time is involved, several values can be recorded simultaneously, giving rise to an ordered array of vectors, which is, in the field of time series, often referred to as multidimensional time series. To name only a few domains, market evolution is described by financial time series, and physiological variables (e.g., electrocardiograms, electroencephalograms)  are recorded simultaneously in medicine, yielding multidimensional time series. Finally, smartphone and GPS sensors data, or character recognition problems, present both spatial and temporal aspects. These high-dimensional datasets open up new theoretical and practical challenges, as both algorithms and statistical methods need to be adapted to their sequential nature. 
    
    Different communities have addressed this problem. First, time series forecasting has been an active area of research in statistics since the 1950s, resulting in several monographs, such as \citet{hamilton1994time}, \citet{box2015time} and \citet{shumway2017time}, to which the reader is referred for overviews of the domain. Time series are considered as realizations of various stochastic processes, such as the famous ARIMA models. Much work in this field has been done on parameter estimation and model selection. These models have been developed for univariate time series but have been extended to the multivariate case \citep{lutkepohl2005new}, with the limitation that they become more complicated and harder to fit. 

    More recently, the field of functional data analysis has extended traditional statistical methods, in particular regression and Principal Component Analysis, to functional inputs. \citet{ramsay2005functional,ferraty2006nonparametric} provide introductions to the area. \citet{kokoszka2017special} give an account of recent advances. In particular, longitudinal functional data analysis is concerned with the analysis of repeated observations, where each observation is a function \citep{greven2011longitudinal,park2015longitudinal}. The data arising from this setting may be considered as a set of vector-valued functions with correlated coordinates, each function corresponding to one subject and each coordinate corresponding to one specific observation. 

    Although these various disciplines  work with sequential data, their goals usually differ. Typically, time series analysis is concerned with predicting future values of one observed function, whereas (longitudinal) functional data analysis usually collects several functions and is then concerned with the prediction of another response variable. However, all these methods rely on strong assumptions on the regularity of the data and need to be adapted to each specific application. Therefore, modern datasets have highlighted their limitations: a lot of choices, in basis functions or model parameters, need to be handcrafted and are valid only on a small-time range. Moreover, these techniques struggle to model multidimensional series, in particular, to incorporate information about interactions between various dimensions. 
    
    On the other side, time series classification has attracted the interest of the data mining community. A broad range of algorithms have been developed, reviewed by \citet{bagnall2017great} in the univariate case. Much attention has been paid to the development of similarity measures adapted to temporal data, a popular baseline being the Dynamic Time Warping metric \citep{berndt1996finding}, combined with a 1-nearest neighbor algorithm. \citet{bagnall2017great} state that this baseline is beaten only by ensemble strategies, which combine different feature mappings. However, a great limitation of these methods is their complexity, as they have difficulty handling large time series. Recently, deep learning seems to be a promising approach and solves some problems mentioned above. For example, \citet{fawaz2019deep} claim that some architectures perform systematically better than previous data mining algorithms. However, deep learning methods are costly in memory and computing power, and often require a lot of training data.
    
    The present article is concerned with a novel approach for sequential learning, called the signature method, and coming from rough path theory. Its main idea is to summarize temporal or sequential inputs by the graded feature set of their iterated integrals, the signature. In rough path theory, functions are referred to as paths, to emphasize their geometrical aspects. Indeed, the importance of iterated integrals had been noticed by geometers in the 60s, as presented in the work of \citet{chen1958integration}. It has been rediscovered by \citet{lyons1998differential} in the context of stochastic analysis and controlled differential equations, and is at the heart of rough path theory. This theory, of which \citet{lyons2007differential} and \citet{friz2010multidimensional} give a recent account, focuses on developing a new notion of paths to make sense of evolving irregular systems. Notably, \citet{hairer2013solving} was awarded a Fields medal in 2014 for its solution to the Kardar-Parisi-Zhang equation built with rough path theory. In this context, it has been shown that the signature provides an accurate summary of a (smooth) path and allows to obtain arbitrarily good linear approximations of continuous functions of paths. Therefore, to learn an output $Y \in \R$, which is an unknown function of a random path $X:[0,1] \to \R^d$, rough path theory suggests that the signature is a relevant tool to describe $X$.
    
    The signature has recently received the attention of the machine learning community and has achieved a series of successful applications. To cite some of them, \citet{yang2016deepwriterid} have achieved state-of-the-art results for handwriting recognition with a recurrent neural network combined with signature features. \citet{graham2013sparse} used the same approach for character recognition, and \citet{lyons2014extracting} coupled Lasso with signature features for financial data streams classification. \citet{kormilitzin2016application} investigated its use for the detection of bipolar disorders, and \citet{yang2017leveraging} for human action recognition. For introductions to the signature method in machine learning, the reader is referred to the work of \citet{levin2013learning} and to \citet{chevyrev2016primer}.
    
    However, despite many promising empirical successes, a lot of questions remain open, both practical and theoretical. In particular, to compute signatures, it is necessary to embed discretely sampled data points into paths. While authors use different approaches, this embedding is only mentioned in some articles, and rarely discussed. Thus, the purpose of this paper is to take a step forward in understanding how signature features should be constructed for machine learning tasks, with a special focus on the embedding step. The article is organized as follows.
    \begin{enumerate}
        \item[$(i)$] In Section \ref{sec:signature_feature_set}, a brief exposition of the signature definition and properties is given, along with a survey of different approaches undertaken in the literature to combine signatures with machine learning algorithms. Datasets used throughout the paper are also presented in Section \ref{sec:datasets}.
        \item[$(ii)$] In Section \ref{sec:embedding}, potential embeddings are reviewed and their predictive performance is compared with an empirical study on 3 real-world datasets. This study indicates that the embedding is a step as crucial as the algorithm choice since it can drastically impact accuracy results. In particular, we find that the lead-lag embedding systematically outperforms other embeddings, consistently over different datasets and algorithms. This finding is reinforced by a simulation study with autoregressive processes in Section \ref{sec:simu_ar}.
        \item[$(iii)$] In Section \ref{sec:sig_domain_and_performance} the choice of signature domain is investigated. Signatures can be computed on any sub-interval of the path definition domain, and it is natural to wonder whether some local information is lost when signatures of the whole path are computed. The section ends by showing that, with a good embedding, the signature combined with a simple algorithm, such as a random forest classifier, obtains results comparable to state-of-the-art approaches in different application areas, while remaining a generic approach and computationally simple.
        \item[$(iv)$] In Section \ref{sec:conclusion} some open questions for future work are discussed.
    \end{enumerate}
    
    These empirical results are based on three recent datasets, in different fields of application. One is a univariate sound recording dataset, called Urban Sound \citep{salamon2014dataset}, whereas the others are multivariate. One has been made available by \citet{quickdrawdata}, and consists of drawing trajectories, while the other is made up of 12 channels recorded from smartphone sensors \citep{malekzadeh2018protecting}. They are each of a different nature and present a variety of lengths, noise levels, and dimensions. In this way, generic and domain-agnostic results are obtained. The code is available at \url{https://github.com/afermanian/embedding_with_signatures}.

    \section{A first glimpse of the signature method}
    \label{sec:signature_feature_set}
    
    \subsection{Definition and main properties}
    \label{sec:sig_definition}
    
    In this subsection, the notion of signature is introduced and some of its important properties are reviewed. The reader is referred to \citet{lyons2007differential} or \citet{friz2010multidimensional} for a more involved mathematical treatment with proofs. Throughout the article, the basic objects are paths, that is, functions from $[0,1] \to \R^d$, where $d \in \N^\ast$. The main assumption is that these paths are of bounded variation, i.e., they have finite length.
    
    \begin{definition}
        Let 
        \begin{align*}
        X : [0,1]& \longrightarrow \R^d \\
        t &\longmapsto (X^1_t, \dots , X^d_t).
        \end{align*}
        The total variation of $X$  is defined by
        \[\|X\|_{1-var} =  \underset{D}{\textnormal{sup}} \sum_{t_i \in D} \|X_{t_i} - X_{t_{i-1}} \| ,\]
        where the supremum is taken over all finite partitions 
        \[D=\big\{(t_0,\ldots,t_k) \, |\, k \geq 1, \, 0=t_0<t_1<\cdots<t_{k-1}<t_k=1 \big\}\] 
        of $[0,1]$, and $\| \cdot \|$ denotes the Euclidean norm on $\R^d$. The path $X$ is said to be of bounded variation if its total variation is finite.
    \end{definition}
    
    The set of bounded variation paths is exactly the set of functions whose first derivatives exist almost everywhere. Being of bounded variation is therefore not a particularly restrictive assumption. It contains, for example, all Lipschitz functions. In particular, if $X$ is continuously differentiable, and $ \dot{X}$ denotes its first derivative with respect to $t$, then
    \[\| X \|_{1-var} = \int_{0}^{1} \| \dot{X}_t \| dt.\]

    The assumption of bounded variation allows us to define Riemann-Stieljes integrals along paths. An exposition of this integration theory is not given here, but the interested reader is referred to \citet{lyons2007differential}. From now on, it is assumed that the integral of a continuous path $Y:[0,1] \to \R^d$ against a path of bounded variation $X:[0,1] \to \R^d$ is well-defined on any $[s,t] \subset [0,1]$, and denoted by
    \[\int_s^t Y_u dX_u = \begin{pmatrix}
    \int_{s}^{t} Y^1_u dX^1_u \\
    \vdots \\
    \int_{s}^t Y^d_u dX^d_u
    \end{pmatrix} \in \R^d, \]
    where $X=(X^1, \dots, X^d)$, and $Y=(Y^1,\dots, Y^d)$. When $X$ is continuously differentiable, this integral is equal to the standard Riemann integral, that is,
    \[\int_s^tY_u dX_u= \int_s^t Y_u \dot{X}_u du. \]
    As an example, assume that $X$ is linear, i.e.,
    \begin{equation}
    \label{eq:linear_path}
    X_t=( X^1_t, \dots, X^d_t)= (a_1 + b_1 t, \dots,  a_d + b_d t), \quad 0 \leq t \leq 1,
    \end{equation}
    where $a_1,\dots ,a_d, b_1, \dots b_d \in \R$. Then
    \[\int_{s}^{t} dX_u = \int_{s}^{t} \dot{X}_u du = \begin{pmatrix}
    \int_{s}^{t} b_1 du \\
    \vdots \\
    \int_{s}^{t} b_d du
    \end{pmatrix} = \begin{pmatrix}
    b_1 (t-s) \\
    \vdots \\
    b_d(t-s)
    \end{pmatrix}.\]
    The formula above is useful since in practice only integrals of linear paths are computed, as discussed later in this subsection. It is now possible to define the signature. 
    
    \begin{definition}
        Let $X:[0,1] \to R^d$ be a path of bounded variation, $I=(i_1,\dots , i_k) \subset \{1,\dots d\}^k$, $k\in \N^\ast$, be a multi-index of length $k$, and $[s,t] \subset [0,1]$ be an interval. The signature coefficient of $X$ corresponding to the index $I$  on $[s,t]$ is defined by
        \begin{equation}
        \label{eq:def_signature_S^I}
        S^I(X)_{[s,t]} =  \idotsint\limits_{s \leq u_1<\dots <u_k \leq t} dX^{i_1}_{u_1} \dots dX^{i_k}_{u_k} = \int_{s}^{t} \bigg(\int_{u_1}^{t} \Big( \int_{u_2}^{t} \dots \int_{u_{k-1}}^{t} dX^{i_k}_{u_k} \Big) dX^{i_{2}}_{u_{2}} \bigg)dX^{i_1}_{u_1} .
        \end{equation}
        $S^I(X)_{[s,t]}$ is then said to be a signature coefficient of order $k$.
    \end{definition}
    The signature of $X$ is the sequence containing all signature coefficients, i.e.,
    \begin{equation*}
    S(X)_{[s,t]}=\big(1,S^{(1)}(X)_{[s,t]},\dots,S^{(d)}(X)_{[s,t]},S^{(1,1)}(X)_{[s,t]},\dots, S^{(i_1,\dots, i_k)} (X)_{[s,t]},\dots  \big) .
    \end{equation*}
    The signature of $X$ truncated at order $K$, denoted by $S_K(X)$, is the sequence containing all signature coefficients of order lower than or equal to $K$, that is
    \begin{equation*}
    S_K(X)_{[s,t]}=\big(1,S^{(1)}(X)_{[s,t]},S^{(2)}(X)_{[s,t]},\dots, S^{\overbrace{(d,\dots ,d)}^K}(X)_{[s,t]} \big) .
    \end{equation*}
    For simplicity, when $[s,t]=[0,1]$, the interval is omitted in the notations, and, e.g., $S_K(X)$ is written instead of $S_K(X)_{[0,1]}$. 

    From these definitions, it follows that the linear interpolation of a (multivariate) time series observed on a finite time horizon will be of bounded variation, and therefore that its signature is well defined. Note that this procedure of mapping a discrete time series into a continuous path is called an embedding, and linear interpolation is only one embedding among others, which will be studied in Section \ref{sec:embedding}. For example, Brownian motion is not of bounded variation but is instead of finite $p$-variation for any $p>2$. However, its signature can still be defined with Itô or Stratonovitch integrals.

    Before giving an explicit calculation of signatures, some comments are in order. First, it is worth noting that, for a path in $\R^d$, there are $d^k$ coefficients of order $k$. The signature truncated at order $K$ is therefore a vector of dimension 
    \begin{equation}
    \label{eq:size_signature}
    \sum_{k=0}^{K} d^k = \frac{d^{K+1}-1}{d-1} \quad \text{if } d \neq 1,
    \end{equation}
    and $K+1$ if $d=1$. Unless otherwise stated, it is assumed that $d\neq 1$, as this is in practice usually the case. Thus, the size of $S_K(X)$ increases exponentially with $K$, and polynomially with $d$---some typical values are presented in Table \ref{tab:truncation_dimensions}. 

    \begin{table}[h]
        \centering
        \begin{tabular}{ l c c c }
        \toprule
            & $d=2$  & $ d=3$   &$ d=6$ \\
            \midrule
            $K=1$& 2 & 3 & 6 \\
            $K=2$& 6 & 12 & 42 \\
            $K=5$ & 62 &363 & 9330 \\
            $K=7$ &254 & 3279 & 335922\\
            \bottomrule
        \end{tabular}
        \caption{Typical sizes of $S_K(X)$ for different values of $K$ and $d$, where $X:[0,1] \to \R^d$.}
        \label{tab:truncation_dimensions}
    \end{table}

    Moreover, the set of coefficients of order $k$ can be seen as an element of the $k$th tensor product of $\R^d$ with itself, denoted by $(\R^d)^{\otimes k}$. For example, the $d$ coefficients of order $1$ can be written as a vector, and the $d^2$ coefficients of order $2$ as a matrix:
    \[\begin{pmatrix}S^{(1)} (X) \\ \vdots \\ S^{(d)} (X) \\ \end{pmatrix} \in \R^d,   \quad \begin{pmatrix}
    S^{(1,1)} (X)& \dots & S^{(1,d)} (X)\\
    \vdots  & & \vdots \\
    S^{(d,1)}(X) & \dots & S^{(d,d)} (X)\\
    \end{pmatrix} \in \R^{d \times d} \approx (\R^d)^{\otimes 2}.\]
    Similarly, coefficients of order 3 can be written as a tensor of order 3, and so on. Then, $S(X)$ can be seen as an element of the tensor algebra
    \[\R \oplus \R^d \oplus (\R^d)^{\otimes 2} \oplus \dots \oplus (\R^d)^{\otimes k} \oplus \cdots. \]
    This structure of the tensor algebra will not be used in the present article but is used to derive properties of the signature \citep{lyons1998differential,friz2010multidimensional,hambly2010uniqueness}. 
    
    It should be noted that due to the ordering in the integration domain in \eqref{eq:def_signature_S^I}, the signature coefficients are not symmetric. For example, $S^{(1,2)}(X)$ is not the same as $S^{(2,1)}(X)$. Finally, \citet{chevyrev2016primer} show how, under certain assumptions, empirical statistical moments can be explicitly recovered from signature coefficients. Typically, the empirical mean can be recovered from signature coefficients of order 1, the variance from coefficients of order 2, and so on. Therefore, the larger the truncation order, the more detailed the information encoded in the signature.
    
    As a toy example, consider the linear path \eqref{eq:linear_path} again, and assume for simplicity that $d=2$: \[X_t= \begin{pmatrix}
    X^1_t \\ X^2_t
    \end{pmatrix} = \begin{pmatrix}
    a_1 + b_1t \\ a_2 + b_2t
    \end{pmatrix}.\] Then, for any $[s,t] \subset [0,1]$ the signature coefficients of order $1$ are
    \begin{align*}
    S^{(1)}(X)_{[s,t]}= \int_{s}^{t} dX^1_u =b_1 (t-s) \quad \text{and} \quad S^{(2)}(X)_{[s,t]}= \int_{s}^{t} dX^2_u =b_2 (t-s) .
    \end{align*}
    The first coefficient of order $2$ is
    \[S^{(1,1)}(X)_{[s,t]}=\int_{s}^{t} \int_{u_1}^{t} dX^1_{u_2} dX^1_{u_1}  =  \int_{s}^{t} \int_{u_1}^{t} b_1^2 du_2 du_1 = b_1^2 \int_{s}^{t} (t-u_1) du_1 = \frac{b_1^2 (t-s)^2}{2}.  \]
    Similarly,
    \[S^{(1,2)}(X)_{[s,t]}=S^{(2,1)}(X) =\frac{b_1 b_2(t-s)^2}{2} \quad  \text{ and } \quad S^{(2,2)}(X)_{[s,t]} =\frac{b_2^2(t-s)^2}{2}.\]
    For any index $I=(i_1,\dots,i_k) \subset \{1,2\}^k$, it is easily obtained that
    \begin{equation}
    \label{eq:formula_sig_linear_path}
    S^{(i_1,\dots ,i_k)}(X)_{[s,t]}=  \idotsint\limits_{s \leq u_1<\dots <u_k \leq t} dX^{i_1}_{u_1} \dots dX^{i_k}_{u_k}= \frac{b_{i_1} \dots b_{i_k}(t-s)^k}{k!}.
    \end{equation}
    
\begin{figure}[h]
        \centering
        \includegraphics[width=.6\textwidth]{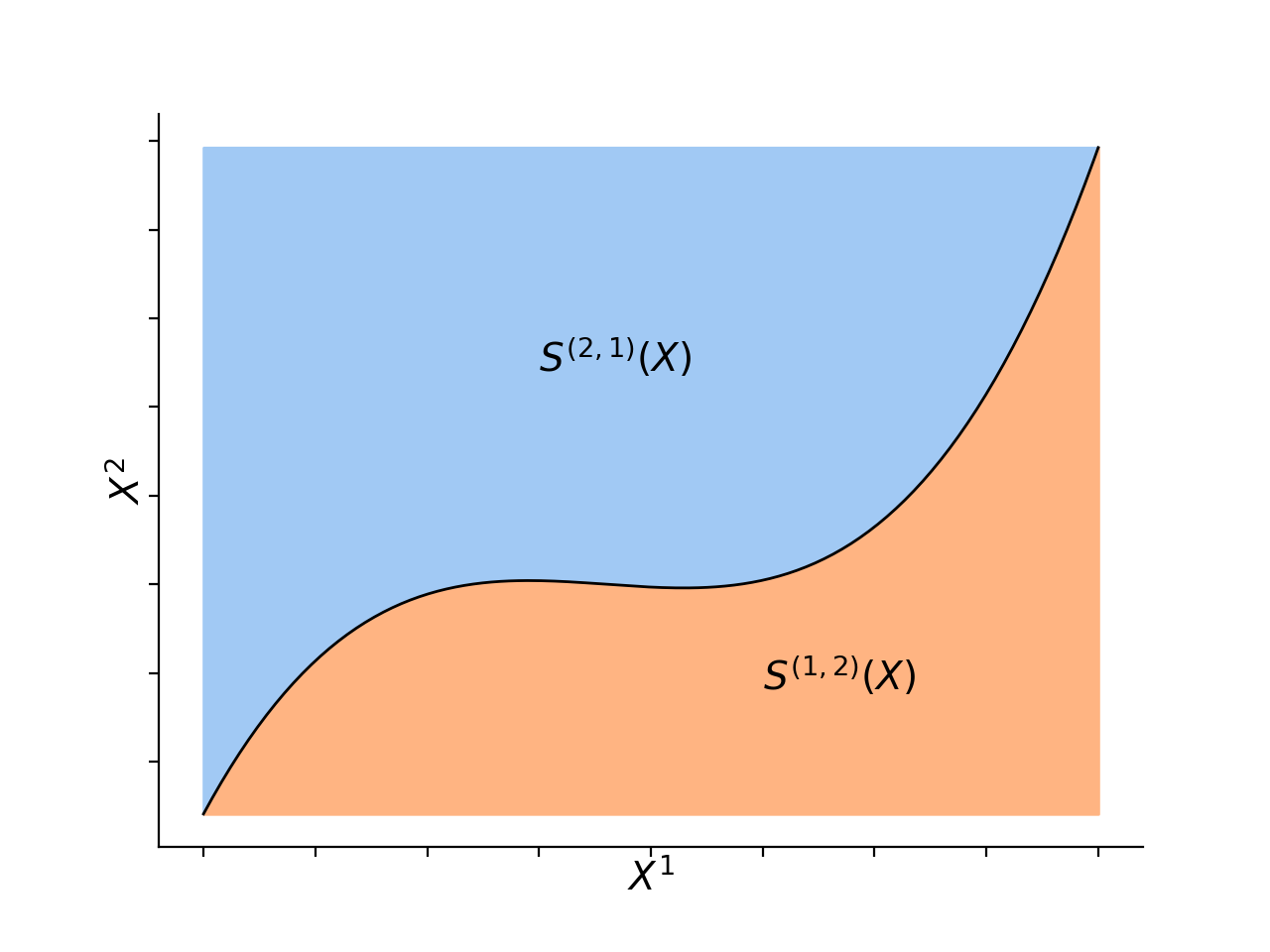}
        \caption{Geometric interpretation of signature coefficients.}
        \label{fig:geom_interpretation}
    \end{figure}

    A crucial feature of the signature is that it encodes geometric properties of the path. Indeed, coefficients of order 2 correspond to some areas outlined by the path, as shown in Figure \ref{fig:geom_interpretation}. For higher orders of truncation, the signature contains information about the joint evolution of tuples of coordinates \citep[][]{yang2017leveraging}. Furthermore, the signature possesses several properties that make it a good statistical summary of paths, as shown in the next four propositions.
    
    \begin{proposition}
        \label{prop:inv_reparam}
        Let $X:[0,1] \to \R^d$ be a path of bounded variation, and $\psi:[0,1] \to [0,1]$ be a non-decreasing surjection. Then, if $\widetilde{X}_t=X_{\psi(t)}$ is the reparametrization of $X$ under $\psi$,
        \[S(\widetilde{X})=S(X) .\]
    \end{proposition}
    
    This proposition is a consequence of the properties of integrals and bounded variation paths \citep[Proposition 7.10]{friz2010multidimensional}. In other words, the signature of a path is the same up to any reasonable time change. There is, therefore, no information about the path parametrization in signature coefficients. However, when relevant for the application, it is possible to include this information by adding the time parametrization as a coordinate of the path. This procedure plays a decisive role in the construction of time embeddings, which will be thoroughly discussed in Section \ref{sec:embedding}. 
    
    A second important property is a condition ensuring the uniqueness of signatures.
    
    \begin{proposition}
        \label{prop:uniqueness}
        If $X$ has at least one monotone coordinate, then $S(X)$ determines $X$ uniquely up to translations.
    \end{proposition}
    
    It should be noticed that having a monotone coordinate is a sufficient condition, but a necessary one can be found in \citet{hambly2010uniqueness}, together with a proof of this proposition. The principal significance of this result is that it provides a practical procedure to guarantee signature uniqueness: it is sufficient to add a monotone coordinate to the path $X$. For example, the time embedding mentioned above will satisfy this condition. 
    
    This result does not provide a practical procedure to reconstruct a path from its signature. However, this is an active area of research \citep{Chang2017,Lyons2017hyperbolic,Lyons2018inverting}. In particular, \cite{Lyons2017hyperbolic} derive an explicit expression of rectilinear paths, defined in Section \ref{sec:def_embedding}, in terms of their signatures; and \cite{Lyons2018inverting} construct, from the signature of a ${\mathcal C}^1$ path, a sequence of piecewise linear approximations converging to the initial path.

    The next proposition reveals that the signature linearizes functions of $X$. We refer the reader to \citet[Theorem 1]{Kiraly2016} for a proof.
    
    \begin{proposition}
        \label{prop:linear_approx}
        Let $D$ be a compact subset of the space of bounded variation paths from $[0,1]$ to $\R^d$ and such that for any $X \in D$, $X_0=0$ and $X$ has at least one monotone coordinate. Let $f : D \rightarrow \R$ be continuous. Then, for every $\varepsilon >0$, there exists $N \in \N$, $w \in \R^N$, such that, for any $X\in D$,
        \[\big|f(X) - \langle w, S(X) \rangle \big| \leq \varepsilon,\]
        where $\langle \cdot,\cdot \rangle$ denotes the Euclidean scalar product on $\R^N$.
    \end{proposition}
    
    This proposition is a consequence of the Stone-Weierstrass theorem. The classical Weierstrass approximation theorem states that every real-valued continuous function on a closed interval can be uniformly approximated by a polynomial function. Similarly, this theorem states that any real-valued continuous function on a compact subset $D$ of bounded variation paths can be uniformly approximated by a linear form on the signature. Linear forms on the signature can, therefore, be thought of as the equivalent of polynomial functions for paths. 
    
    Chen's theorem \citep{chen1958integration} now provides a formula to compute recursively the signature of a concatenation of paths. Let $X:[s,t] \rightarrow \R^d$ and $Y : [t,u] \rightarrow \R^d$ be two paths, $0 \leq s < t < u \leq 1$. Then, the concatenation of $X$ and $Y$, denoted by $X \ast Y$, is defined as the path from $[s,u]$ to $\R^d$ such that, for any $v \in [s,u]$,
    \[(X \ast Y)_v = \begin{cases}
    X_v &\textnormal{ if } v \in [s,t], \\
    X_t + Y_{v} - Y_t & \textnormal{ if } v \in [t,u].\\
    \end{cases} 
    \]
    
    \begin{proposition}[Chen]
        \label{prop:chen}
        Let $X:[s,t] \rightarrow \R^d$ and $Y : [t,u] \rightarrow \R^d$  be two paths with bounded variation. Then, for any multi-index $(i_1,\dots ,i_k) \subset \{1,\dots ,d\}^k$,
        \begin{equation}
        \label{eq:chen_identity}
        S^{(i_1,\dots,i_k)}(X \ast Y) =  \sum_{\ell=0}^{k} S^{(i_1,\dots, i_\ell)}(X) \cdot S^{(i_{\ell+1}, \dots, i_k)}(Y).
        \end{equation}
    \end{proposition}
    
    This proposition is an immediate consequence of the linearity property of integrals \citep[Theorem 2.9]{lyons2007differential}. However, it is essential for the explicit calculation of signatures. Indeed, in practice, $X$ is observed at a finite number of times and becomes by interpolation a continuous piecewise linear path. To compute its signature, it is then sufficient to iterate the following two steps:
    \begin{enumerate}
        \item Compute with equation \eqref{eq:formula_sig_linear_path} the signature of a linear section of the path.
        \item Concatenate it to the other pieces with Chen's formula \eqref{eq:chen_identity}.
    \end{enumerate}
    This procedure is implemented in the Python library \pkg{iisignature}   \citep{reizenstein2018iisignature}. Thus, for a sample consisting of $\ell$ points in $\R^d$, if the path formed by their linear interpolation is considered, the computation of the path signature truncated at level $K$ takes $O(\ell d^K)$ operations. The complexity is therefore linear in the number of sampled points but exponential in the truncation order $K$. Notice that the size of the signature vector is also exponential in the truncation order $K$, as shown in Table \ref{tab:truncation_dimensions}. Therefore, in applications, $K$ has to remain small, typically of the order less than 10. 

    We conclude this section by giving some insights into the behavior of the expected value of the signature. Let $(\Omega, {\mathcal{F}}, \prob)$ be a probability space and $X$ a $\R^d$-valued stochastic process defined on $[0,1]$. \citet{chevyrev2016characteristic} have shown that, under some assumptions, $\esp[S(X)]$ characterizes the law of $X$. In other words, if $X$ and $Y$ are two stochastic processes such that $\esp[S(X)]=\esp[S(Y)]$, then $X$ and $Y$ have the same distribution. The assumptions of this result have been relaxed by \citet{chevyrev2018signature}, who only need to assume that the signature is well-defined to prove a similar result for a particular renormalization of the signature. It is instructive to compare this property to the case of random variables. Indeed, recall that if $X$ is a real-valued random variable, then its moment-generating function, defined by $t \mapsto \esp[e^{tX}]$ characterizes the law of $X$. Now if $X$ is a stochastic process, its expected signature has the same property and should, therefore, be thought of as a generalization of the moment-generating function of a process. This interpretation of the signature is particularly clear in the case where $d=1$: let $X:[0,1] \mapsto \R$, then 
    \[\esp[S(X)] =\Big(1, \esp[X_1-X_0], \frac{\esp[(X_1-X_0)^2]}{2!}, \dots,  \frac{\esp[(X_1-X_0)^k]}{k!}, \dots \Big),\]
    and the signature corresponds exactly to an infinite sequence of the moments of the path.
    
    \subsection{Signature and machine learning }
    \label{sec:signature_ml_review}
    
    Now that the signature and its properties have been presented, we focus on its use in machine learning. In a statistical context, our goal is to understand the relationship between a random input path $X:[0,1] \to \R^d$ and a random output $Y \in \R$. In a classical setting, we would be given a set of independent and identically distributed (i.i.d.) observations $\big\{ (X_1,Y_1), \dots, (X_n,Y_n) \big\}$, drawn from $(X,Y)$. However, in applications, a realization $X_i$ is observed only at a discrete set of times $0 \leq t_1 < \dots < t_{\ell_i} \leq 1$, $\ell_i \in \N^\ast$. Therefore, we are given an i.i.d. sample  $ \big\{ ( \mathbf{x_1},Y_1), \dots, (\mathbf{x_n},Y_n) \big\}$, where $\mathbf{x_i}$ takes the form of a matrix:
    \begin{equation}
    \label{eq:def_input_matrix}
    \mathbf{x_i} = \begin{pmatrix}
    x^1_{i,1} & \dots& x^1_{i,\ell_i} \\
    \vdots & & \vdots \\
    x^d_{i,1} & \dots& x^d_{i,\ell_i} \\
    \end{pmatrix} \in \R^{d \times \ell_i}.
    \end{equation}
    In this notation, $x^k_{i,j}$ denotes the $k$th coordinate of the $i$th sample observed at time $t_j$. 

    It is worth clarifying the terminology used and how it relates to other disciplines. In time series analysis, $\mathbf{x_i}$ would be a time series evolving in $\R^d$ and observed \textit{at} $\ell_i$ time points. The output $Y$ would be a future step of the time series, for example, if $d=1$ then $Y_i = x^1_{i,\ell_i +1}.$ If $d=1$, then $\mathbf{x_i}$ is a univariate time series, whereas if $d>1$, then it is called a multidimensional, multivariate, vector-valued or multiple time series. The parameter $\ell_i$ would be the series length or the time horizon. If all time series have the same length $\ell$, then notation simplifies as $\mathbf{x_i} \in \R^{d \times \ell}$. On the other hand, in functional data analysis, $X_i$ would be called functional data, a functional observation, or a curve, $Y_i$ would be a scalar response, and $\ell_i$ would be the number of measurements. If this functional data was longitudinal, each $x_{i,\cdot}^k$ would be a functional observation or a profile of the subject $i$, and $d$ would be the number of repeated measurements. Finally, in time series classification, $\mathbf{x_i}$ is a time series and $Y_i$ the label of its class. Borrowing from the machine learning vocabulary, we will also refer to $\mathbf{x_i}$ as the raw data or the input data, and to $Y_i$ as the output or the response.

    The assumption that $Y_i$ is a real number excludes several situations from our study. For example, the goal of functional longitudinal data analysis is usually the prediction of the next functional profile, which does not fall within our setting. Similarly, prediction of functional responses, which are a topic of interest in functional data analysis, or of multiple time points in time series analysis, are not considered. 

    Finally, it is worth noting the dependence of the length $\ell_i$ on $i$. In other words, each observation may have a different length. The signature dimension being independent of the number of sampled points, representing time series by their signature naturally handles inputs of various lengths, whereas traditional methods often require them to be normalized to a fixed length. Moreover, no assumption is made on the sampling intervals $t_1, \dots, t_{\ell_i}$, which can therefore be irregularly spaced and vary from one sample to another. To sum up, the signature method is appropriate for learning with discretely sampled multidimensional time series, possibly of different lengths and irregularly sampled.
    
\begin{figure}[h]
        \centering
        \includegraphics[width=0.8\linewidth]{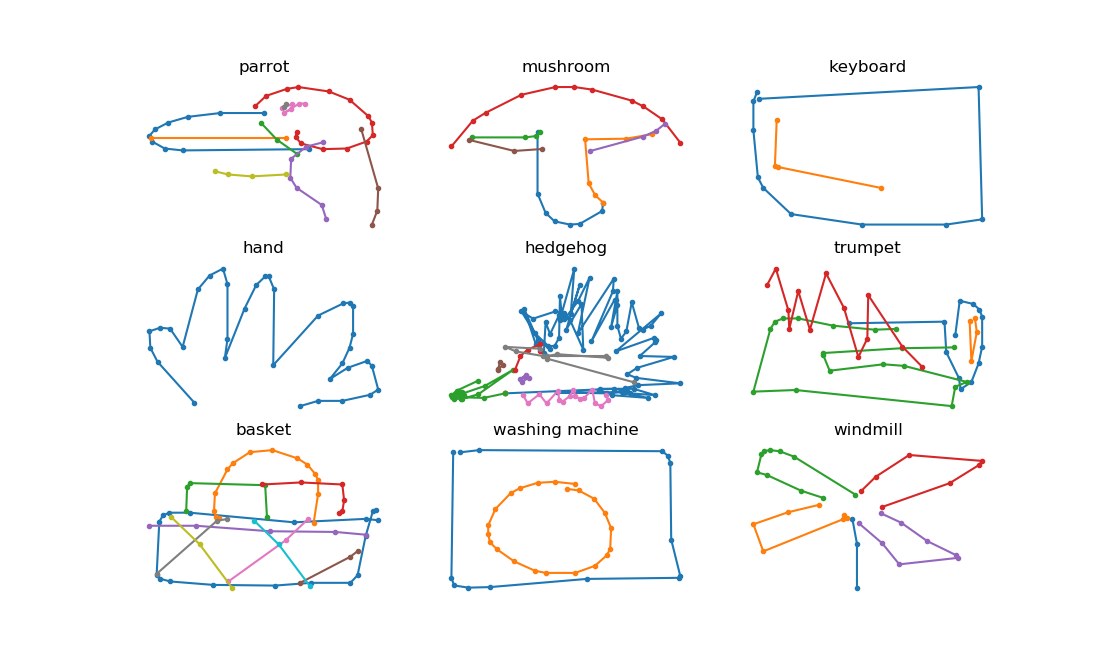}
        \caption{$9$ drawings from the Quick, Draw!~dataset}
        \label{fig:examples_drawings}
    \end{figure}

    As an example, consider the Google dataset Quick, Draw! \citep{quickdrawdata}. It consists of the pen trajectories of millions of drawings, divided into 340 classes. Some examples are shown in Figure \ref{fig:examples_drawings}. In this case, the $y_i$ are discrete labels of the drawing's class, and the $\mathbf{x_i}$ are matrices of pen coordinates. In this example, $d=2$ and $p_i$ varies for each drawing but is typically in the order of a few dozen points.
    
    As discussed in the introduction, to use signature features, one needs to embed the observations $\mathbf{x_i}$ into paths of bounded variation $X_i:[0,1] \to \R^d$. This step, which also consists of adding other coordinates, such as time, will be thoroughly discussed in Section \ref{sec:embedding}. Assume for the moment that a set of embeddings $X_i$, $1 \leq i \leq n$, are given. When an embedding has been chosen, one can compute signature features truncated at a certain order $K$ and use them in combination with a learning algorithm. The choice of $K$ corresponds to a classical bias-variance tradeoff: the larger $K$, the larger the feature set. Therefore, it should be selected in a data-driven way, for example with cross-validation. The procedure can be summarized as follows:
    \[\text{Raw data } \rightarrow \text{Embedding}  \rightarrow \text{Signature features} \rightarrow \text{Algorithm} .\]
    
    The literature on the combination of signature features with learning algorithms can be divided into three groups. These groups correspond to the nature of the algorithm's input: it is either a vector, a sequence, or an image. In the context of deep learning, this division matches the different classes of neural network architectures: feedforward, recurrent and convolutional networks. The latter only deals with input paths in $\R^2$, with applications such as characters or handwriting recognition. 
    
    The first approach is to compute the signature of $X$ on its whole domain, that is, on $[0,1]$. In this way, the time-dependent input $X$ is mapped into a time-independent finite set of coefficients, that is then fed into a predictive algorithm, typically a feedforward neural network. Any time-independent additional covariates may be added to this feature set. This strategy is implemented by \citet{yang2017leveraging} for skeleton-based human action recognition. From a sequence of human joints' positions, the authors construct a high dimensional vector of signature coefficients, which is then the input of a small dense network. \citet{lyons2014extracting,lyons2014feature} also apply this method to financial time series, combining it with Lasso and ordinary least squares regression. 
    
    \begin{figure}[h]
        \centering
        \includegraphics[width=\textwidth]{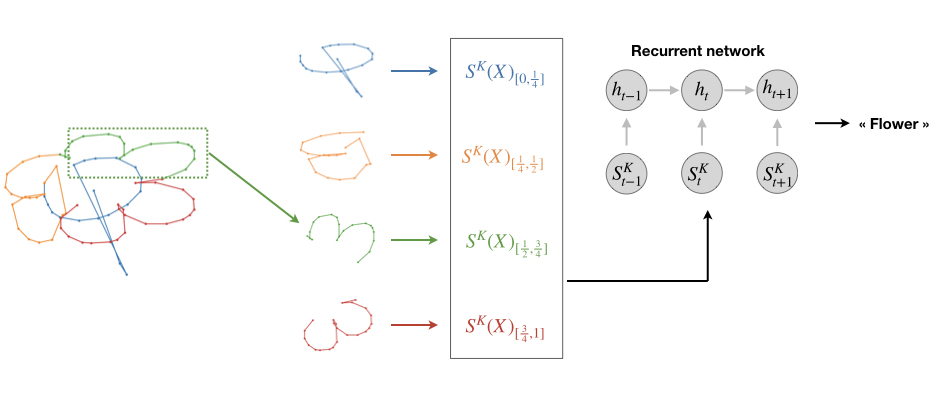}
        \caption{Signature and recurrent neural network.}
        \label{fig:diagram_signature_rnn}
    \end{figure}
    
    A second family of methods consists in describing the input path by a sequence of signature coefficients. There are several variants of this approach, but the one of \citet{wilson2018path} is presented here in detail for its simplicity and representativeness. To create a signature sequence, the time interval $[0,1]$ is divided into a dyadic partition
    \begin{equation}
    \label{eq:dyadic_partition}
    0 \leq 2^{-q} < \dots < j2^{-q} < \dots < (2^q-1) 2^{-q} \leq 1,
    \end{equation}
    where $q \in \N$. By computing the signature truncated at order $K$ on every dyadic interval ${[j 2^{-q}, (j+1) 2^{-q}]}$, $0 \leq j < 2^q$, a sequence of $2^q$ signature vectors is obtained, each of dimension ${(d^{K+1}-1)/(d-1)}$. This sequence is typically fed into a recurrent network, as illustrated in Figure \ref{fig:diagram_signature_rnn}. In this case, the whole approach boils down to transforming the original sequential data into another sequence of signature coefficients. Such a procedure may be surprising, as the original data could have been itself the input of a recurrent network, instead of being mapped into a new signature sequence. However, \citet{lai2017online}, \citet{liu2017ps} and  \citet{wilson2018path} show the superiority of this type of approach for several tasks such as writer and forgeries recognition.

    \begin{figure}[h]
        \centering
        \includegraphics[width=0.95\textwidth]{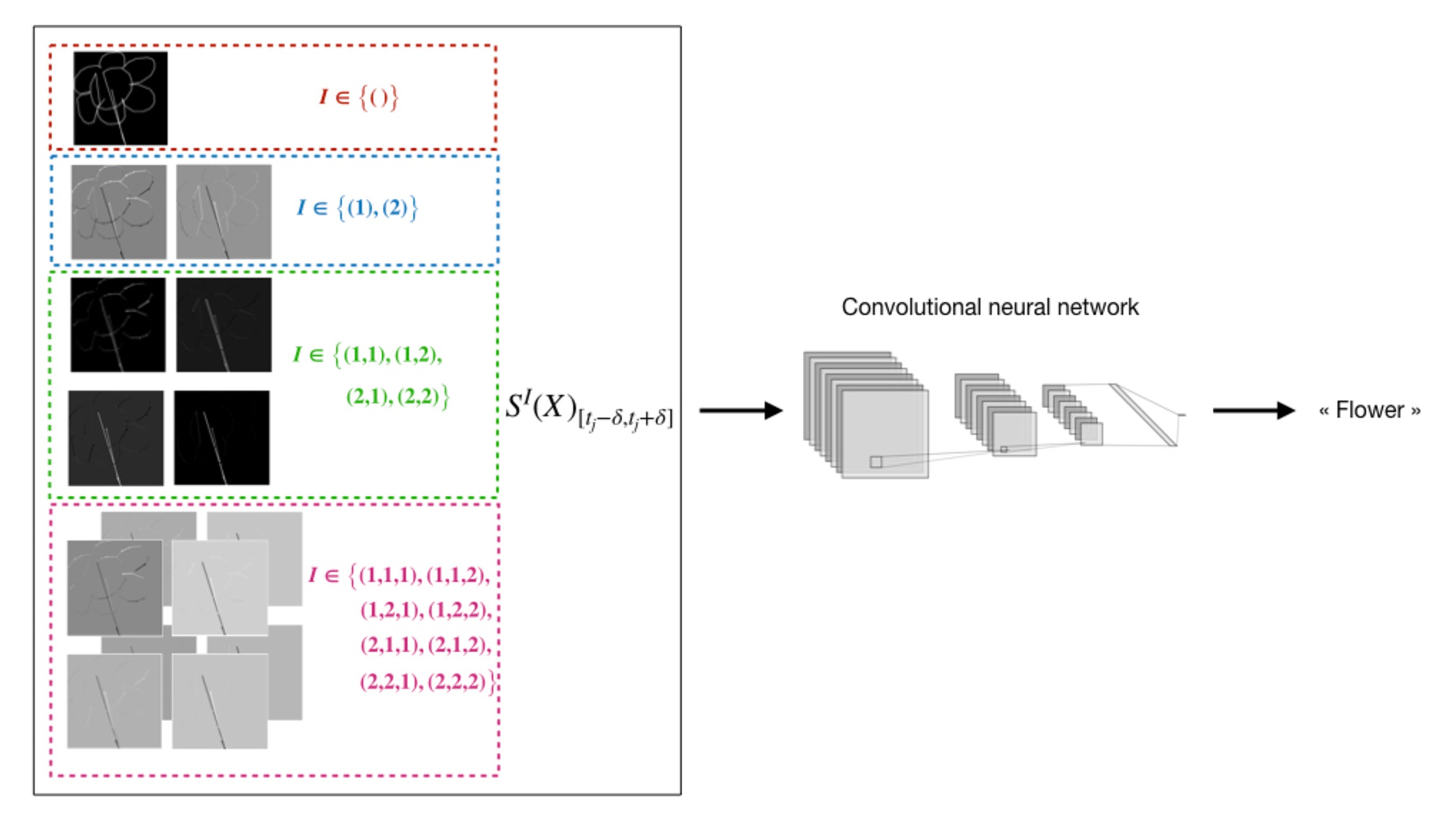}
        \caption{Signature and convolutional neural network.}
        \label{fig:diagram_signature_cnn}
    \end{figure}

    Finally, a third group of authors has taken these ideas further and created images of signature coefficients. Their rationale is to mix the temporal and pictorial aspects of the data. Indeed, assuming that the input path is a trajectory in $\R^2$, it can be turned into an image by forgetting its temporal aspect and setting the pixel values to 1 along the trajectory, and 0 elsewhere. Then, starting from this representation, a bunch of images is created, such that each image corresponds to a signature coefficient, as shown in Figure \ref{fig:diagram_signature_cnn}. This can be done in various ways. However, the general idea is to consider a sliding window following the path and to set the pixel value at the center of the window to be equal to the signature coefficient computed over the window. If the signature is truncated at order $K$, this yields $2^{K+1}-1$ sparse grey pictures, which can then be the input of a convolutional neural network. \citet{graham2013sparse} and  \citet{yang2015chinese,yang2016deepwriterid} have obtained significant accuracy improvements for character recognition and writer identification with this approach.
    
    To sum up, the signature may be used in various ways, and for different applications. Several points of view coexist, and none of them has shown to be systematically better. In particular, on the one hand, signatures may be used to remove temporal aspects and to reduce the dimension of the problem, whereas, on the other hand, they may do the opposite and increase the dimension of the algorithm's input. Moreover, they are combined with various learning algorithms and it may be hard to distinguish the properties of the signature from those of the algorithms. Nevertheless, all these methods assume that discrete data points have been embedded into actual continuous paths. As will be seen in Section \ref{sec:embedding}, the choice of path is crucial. Therefore, we describe in the next section the datasets used throughout the article to understand their underlying structure and find suitable embeddings. 
    
    \section{Datasets}
    \label{sec:datasets}
    
    The datasets used in this article have been chosen to cover a broad range of applications while being recent and challenging in various ways. Moreover, they present a variety of sampling frequencies and dimensions. They illustrate therefore different potential embeddings. 
    
    First, the Quick, Draw!~dataset \citep{quickdrawdata}, which was already discussed in Section \ref{sec:signature_ml_review}, and illustrated in Figure \ref{fig:examples_drawings}, is a public Google dataset. It consists of $50$ million drawings, each drawing being a sequence of time-stamped pen stroke trajectories, divided into 340 categories. It takes approximately 7 gigabytes of hard disk space and is, therefore, a particularly large dataset. To compute the signature of every sample, it would thus be necessary to design a specific architecture, which cannot be implemented on a standard laptop computer. However, the goal of this study is not to achieve the best possible performance, but to understand embedding properties. Moreover, the experiments should be easily reproducible without requiring much computational ressources. Therefore, only a subset of the data is used: 68 000 training samples in Sections \ref{sec:path_embedding_study} and \ref{sec:subpaths_signature}, 12 million in Section \ref{sec:performance_signature}. 
    
    Let us describe more precisely the data format. When an object is drawn, two pieces of information are recorded: pen positions, sampled at different times, and pen jumps. Therefore, one drawing consists of a set of strokes, one stroke being a segment of the drawing between two pen jumps, represented with different colors in Figure \ref{fig:examples_drawings}. As each stroke can be of different length, if $\ell_{i,s}$ is the number of points in the $s$th stroke of drawing number $i$, and if this drawing has $S_i$ strokes, then one drawing consists of $S_i$ tables of sizes $2\times \ell_{i,1}, \dots, 2\times \ell_{i,S_i}$, where the factor 2 corresponds to the plane $\R^2$. For example, in Figure \ref{fig:examples_drawings}, the windmill drawing has $S_i=5$ strokes: the first stroke is the blue one with $3$ points ($\ell_{i,1}=3$), the second one the orange with $\ell_{i,2}=5$ points, and so on. The data has been preprocessed by Google, resulting in the so-called  ``simplified drawing files''. The reader is referred to \citet{quickdrawdata} for a complete description of the preprocessing steps. Finally, each sample $i$ can be encoded under the following compact form:

    \begin{equation}
    \label{eq:quick_draw_matrix}
    \mathbf{x_i} = \begin{pmatrix}
    x^1_{i,1} & \dots & x^1_{i,\ell_{i,1}} &  \dots& x^1_{i,\ell_{i,1}+ \dots + \ell_{i,S_i-1} +1} &\dots& x^1_{i,\ell_{i,1}+ \dots + \ell_{i,S_i}} \\
    x^2_{i,1} & \dots & x^2_{i,\ell_{i,1}} &  \dots&x^2_{i,\ell_{i,1}+ \dots + \ell_{i,S_i-1} +1} &\dots& x^2_{i,\ell_{i,1}+ \dots + \ell_{i,S_i}} \\
    1 & \dots & 1 &  \dots & S_i & \dots & S_i\\
    \end{pmatrix} \in \R^{3 \times \ell_i},
    \end{equation}
    where $(x^1_{i,j},x^2_{i,j})$ are the coordinates of the $j$th point of the drawing number $i$, and there is a pen jump when the last row of $\mathbf{x_i}$ increases by 1. As in \eqref{eq:def_input_matrix}, $\ell_i=\ell_{i,1} + \dots + \ell_{i,S_i}$ denotes the total number of points of drawing $i$. 

    \begin{figure}[h]
        \centering
        \includegraphics[width=\textwidth]{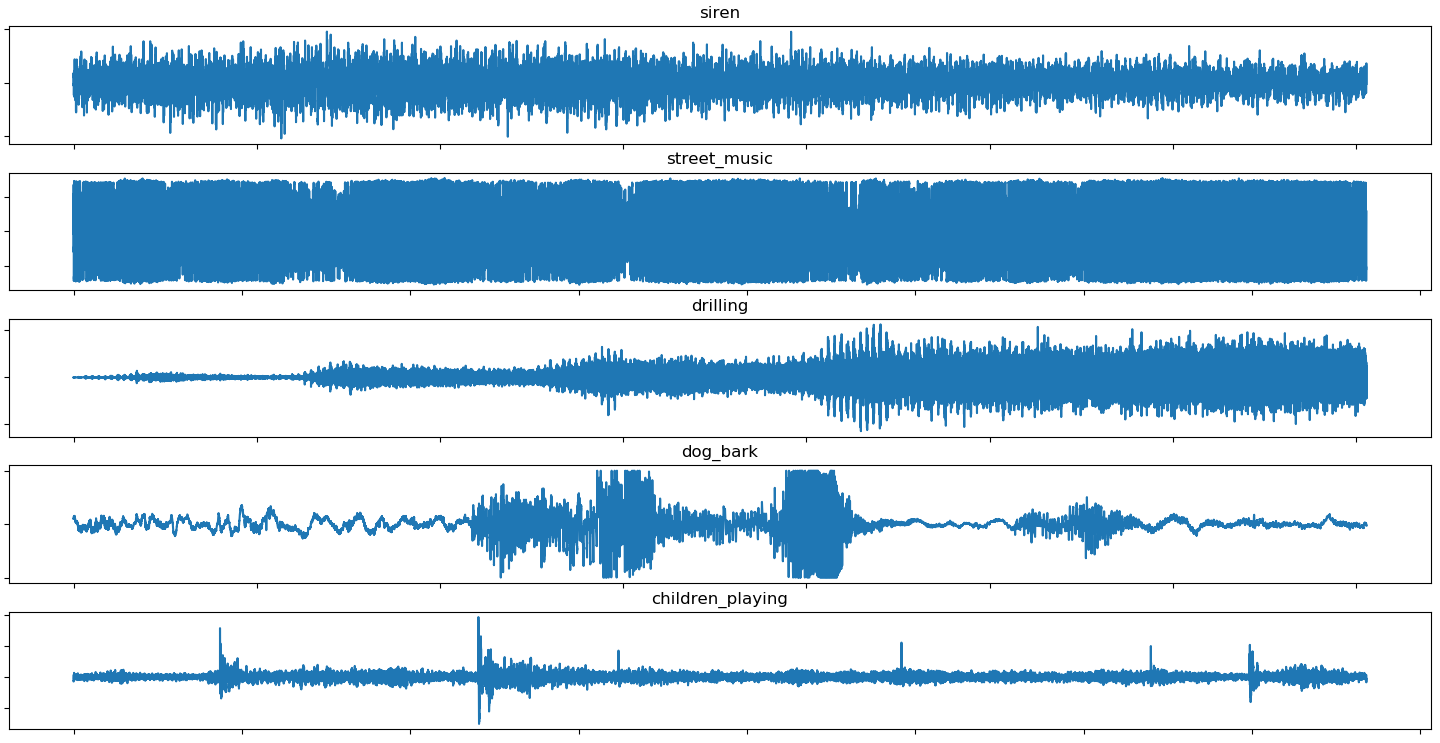}
        \caption{$5$ samples from the Urban sound dataset}
        \label{fig:examples_urban_sound}
    \end{figure}
    
    The second dataset is the Urban Sound dataset \citep{salamon2014dataset}. It consists of sound recordings, divided into 10 classes: car horn, dog barking, air conditioner, children playing, drilling, engine idling, gunshot, jackhammer, siren, and street music. It contains both mono and stereo recordings, so some samples take values in $\R$, and some in $\R^2$. By averaging the two channels of stereo sounds, the data has been normalized to mono recordings, so that each sample is a one-dimensional time series: $\mathbf{x_i} \in \R^{1 \times \ell_i}$. This yields a collection of 5 435 time series of various lengths. On average, they are sampled at approximately 170 000 points, which makes them long time series, typically hard to model along the whole time range. Figure \ref{fig:examples_urban_sound} depicts some examples of these noisy time series. 
   
    \begin{figure}[h]
        \centering
        \includegraphics[width=\textwidth]{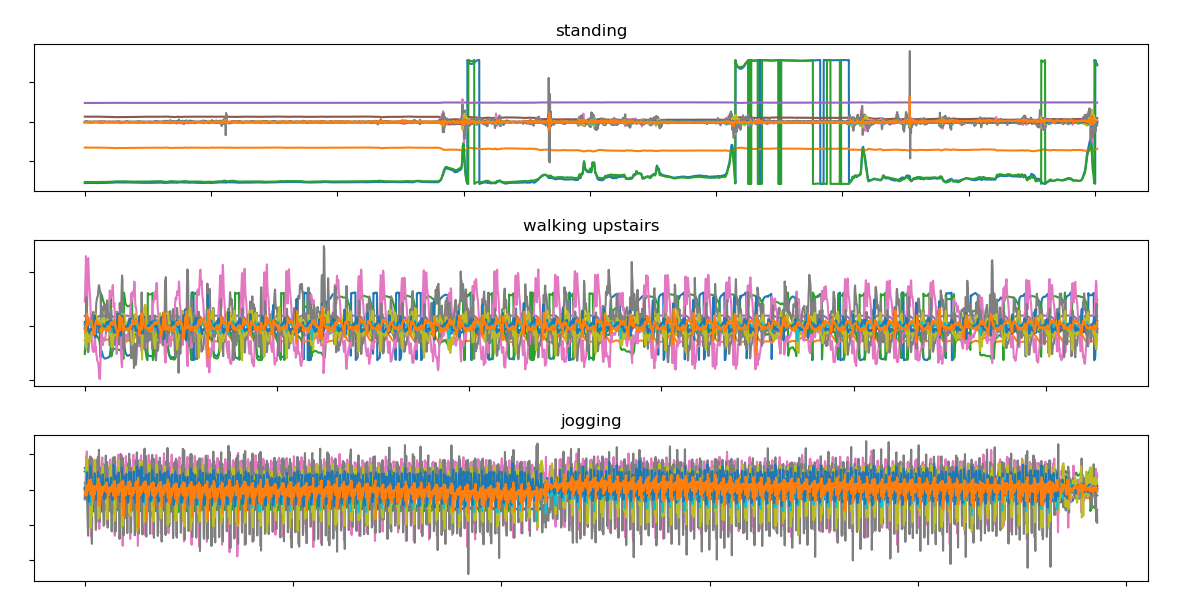}
        \caption{$3$ samples from the Motion sense dataset}
        \label{fig:examples_motion_sense}
    \end{figure} 
    
    Finally, the MotionSense dataset is composed of smartphone sensory data generated by accelerometer and gyroscope sensors \citep{malekzadeh2018protecting}. This data has been recorded while some participants performed an activity among walking upstairs and downstairs, walking, jogging, sitting, and standing. In total, there are 10 classes and 360 recordings, which correspond to 24 participants performing 15 different trials. During each trial, 12 variables are measured: 3 directions for attitude, gravity, user acceleration, and rotation rate, respectively. Information about the participants is provided but we focus on the task of recognizing the activity performed from the multidimensional time series formed by sensors data. In Figure \ref{fig:examples_motion_sense}, three samples are shown, and, for each of them, curves of different colors correspond to the various quantities measured by sensors. Therefore, every sample can be written as $\mathbf{x_i} \in \R^{12 \times \ell_i}$. It is clear from Figure \ref{fig:examples_motion_sense} that these series are noisy and highly dimensional.  They are shorter than the Urban Sound's ones, with an average of approximately 4 000 time steps.

    Finally, for all datasets, the response $Y$ is a class label, that is $Y \in \{1, \dots, C\}$, where $C$ is the number of classes. Table \ref{tab:dataset_summary} summarizes some characteristics of these three datasets. As each sample may have a different length, the data average length is recorded, defined by $\overline{\ell} = \frac{1}{n} \sum_{i=1}^n \ell_i$. These datasets illustrate the diversity of problems in sequential learning, where time appears in different ways.

    \begin{table}[!h]
        \centering
        \begin{tabular}{lc c c }
            \toprule
            & Quick, Draw!~& Urban Sound & Motion Sense \\
            \midrule
            Number of classes $C$ & 340 &10 & 6 \\
            Dimension $d$& 2 & 1 & 12 \\
            Average length $\overline{\ell}$ & 44 &171 135 &3 924 \\
            Training set size $n$ &68 000 & 4 435 & 300\\
            Validation set size $n_{\text{val}}$ &6 800 & 500& 30 \\
            Test set size $n_{\text{test}}$ &6 800 & 500& 30 \\
            \bottomrule
        \end{tabular}
        \caption{Datasets summary}
        \label{tab:dataset_summary}
    \end{table}   
    
    \section{The embedding}
    \label{sec:embedding}
    
    In practice, a matrix of observations $\mathbf{x_i} \in \R^{d \times \ell_i}$ is given, written in \eqref{eq:def_input_matrix}, where columns correspond to points in $\R^d$ sampled at times $0 \leq t_1< \dots < t_{\ell_i} \leq 1$. As explained in Section \ref{sec:signature_ml_review}, the goal is to construct a continuous path $X_i : [0,1] \to \R^d$ from the matrix $\mathbf{x_i}$. Therefore, an interpolation method needs to be chosen, but, to ensure some properties such as signature uniqueness (Proposition \ref{prop:uniqueness}), new coordinates may be added to the path, which increases the dimension $d$ of the embedding space. When not hidden, the embedding is generally only mentioned in the literature, without in-depth discussions. Therefore, the purpose is not only to compare embeddings' performance, but also to give a first systematic survey of their use in the context of learning with signatures.
    
    \subsection{Definition and review of potential embeddings}
    \label{sec:def_embedding}
    
    First, different embeddings are reviewed while adapting them to the Quick, Draw! dataset for illustrative purposes. The extension to other datasets follows immediately. All embeddings considered here are continuous piecewise linear, but their difference lies in the way this interpolation is performed. From a computational point of view, signatures of continuous piecewise linear paths can be computed with the library \pkg{iisignature}, as mentioned in Subsection \ref{sec:sig_definition}. From now on, consider a sample $\mathbf{x} \in \R^{3 \times p}$, which can be written as the matrix \eqref{eq:quick_draw_matrix} where the index $i$ has been removed to simplify notations.
    
    \paragraph{Linear path}
    
\begin{figure}[h]
        \centering
        \begin{subfigure}[b]{0.33\textwidth}
            \includegraphics[width=\textwidth]{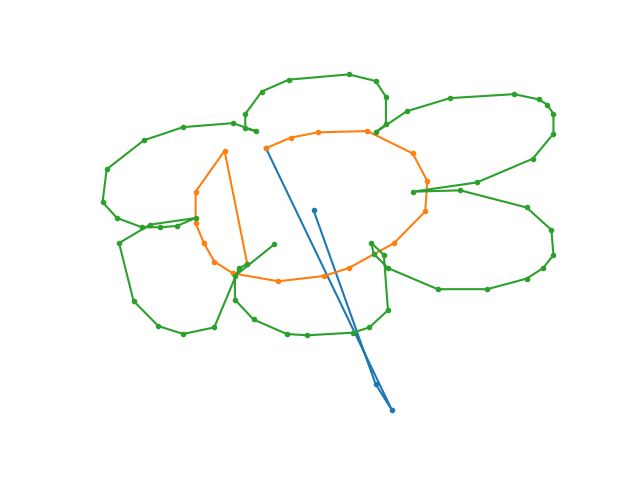}
            \caption{Linear path.}
            \label{fig:flower_raw}
        \end{subfigure}%
        ~
        \begin{subfigure}[b]{0.33\textwidth}
            \includegraphics[width=\textwidth]{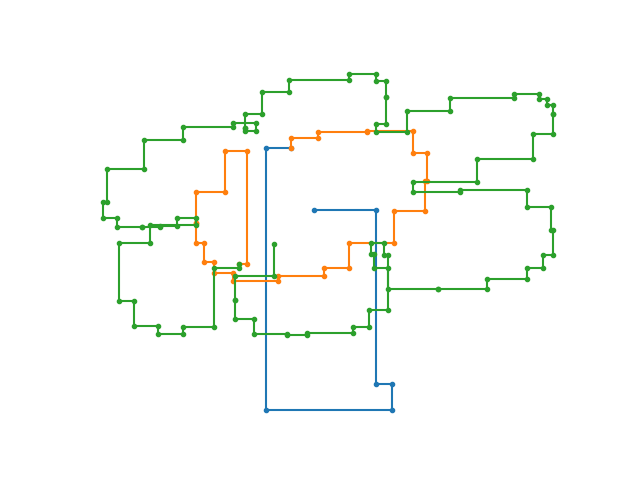}
            \caption{Rectilinear path.}
            \label{fig:flower_rectilinear}
        \end{subfigure}%
        ~
        \begin{subfigure}[b]{0.33\textwidth}
            \includegraphics[width=\textwidth]{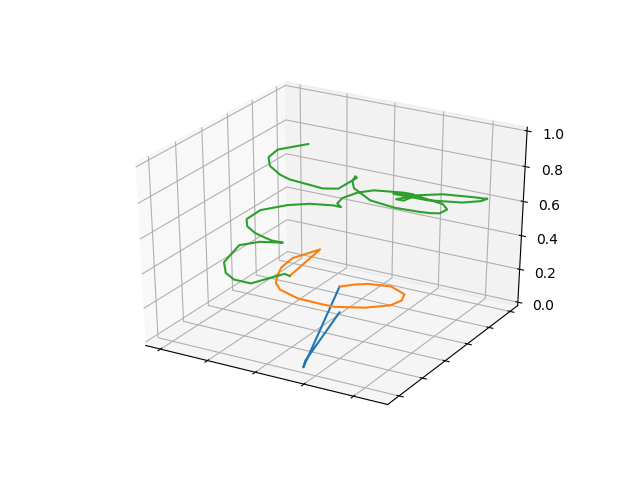}
            \caption{Time path.}
            \label{fig:flower_time_3d_path}
        \end{subfigure}%
        
        \begin{subfigure}[b]{0.33\textwidth}
            \includegraphics[width=\textwidth]{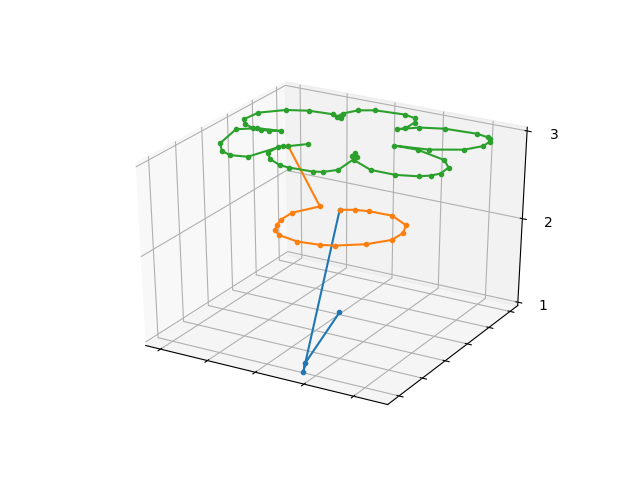}
            \caption{Stroke path, version 1.}
            \label{fig:flower_stroke_3d_path}
        \end{subfigure}%
        ~
        \begin{subfigure}[b]{0.33\textwidth}
            \includegraphics[width=\textwidth]{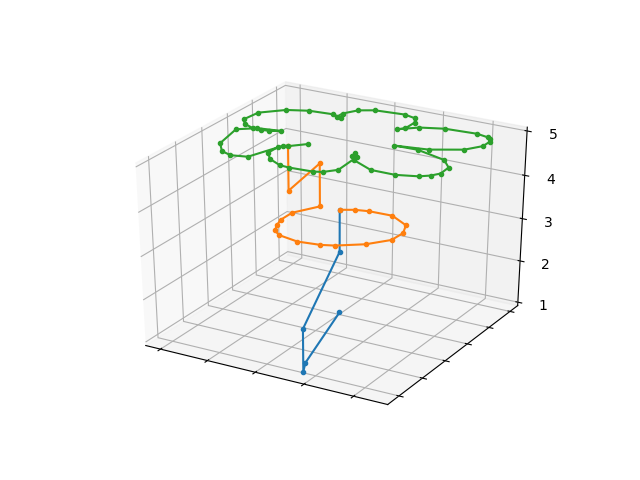}
            \caption{Stroke path, version 2.}
            \label{fig:flower_stroke_3d_path_2}
        \end{subfigure}%
        ~
        \begin{subfigure}[b]{0.33\textwidth}
            \includegraphics[width=\textwidth]{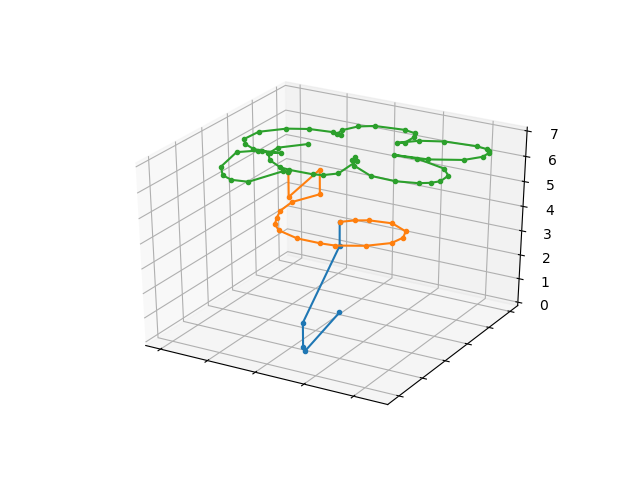}
            \caption{Stroke path, version 3.}
            \label{fig:flower_3d_path_time_stroke}
        \end{subfigure}
        
        \caption{Different embeddings of a Quick, Draw!~sample. Each stroke is plotted with a different color only for the sake of illustration.}
        \label{fig:flower_paths_embeddings}
    \end{figure}

    A first natural choice is to interpolate data points linearly, that is to connect each consecutive points by a straight line. Note that for the Quick Draw!~data, information about pen jumps is then lost. Thus, for a particular sample, if $\ell$ positions of the pen $[(x^1_1,x^2_1), \dots ,(x^1_\ell,x^2_\ell) ]$ are given, the piecewise linear path $X:[0,1] \to \R^2$ is defined as the path equal to $(x^1_j,x^2_j)$ at $t_j$, where $0=t_1< t_2 < \dots < t_\ell=1$ is a partition of $[0,1]$ into $\ell$ points. This yields a two-dimensional continuous path with coordinates $(X^1_t, X^2_t)$. This path, represented in Figure \ref{fig:flower_raw}, is the most often used in the literature, for example by \citet{graham2013sparse}, \citet{lai2017online}, or \citet{yang2016deepwriterid}. 
    
    \paragraph{Rectilinear path}
    
    Another interpolation method is often used in the literature \citep[][]{chevyrev2016primer,kormilitzin2016application} and referred to as an ``axis path" or ``rectilinear path''. It is also piecewise linear but each linear section is parallel to an axis. In other words, to move from one point $(x_j^1,x_j^2)$ to another point $(x_{j+1}^1,x_{j+1}^2)$, a first linear segment goes from $(x_j^1,x_j^2)$ to  $(x_{j+1}^1,x_j^2)$, parallel to the x-axis, and a second segment from $(x_{j+1}^1,x_j^2)$ to $(x_{j+1}^1,x_{j+1}^2)$, parallel to the y-axis. This path is depicted in Figure \ref{fig:flower_rectilinear}. A crucial aspect of this path is that there exists a simple way to reconstruct it from its signature features \citep{Lyons2017hyperbolic}. Note that for unidimensional data, such as the Urban Sound dataset, the linear and rectilinear interpolations are identical.

    \paragraph{Time path}
    
    The third approach builds upon the linear path and enriches it by adding a monotone coordinate. This ensures the uniqueness of the signature, as stated in Proposition \ref{prop:uniqueness}. It usually corresponds to adding the time parametrization as a coordinate of the path, as is done by \citet{yang2017leveraging}. Therefore, if $ t \mapsto (X^1_t,X^2_t)$ is the linear path described above, which is piecewise linear, the time embedding is the $3$-dimensional path $t \mapsto (X^1_t,X^2_t,t)$, shown in Figure \ref{fig:flower_time_3d_path}.

    \paragraph{Lead-lag path}
    
\begin{figure}[h]
        \centering
        \includegraphics[width=0.5\textwidth]{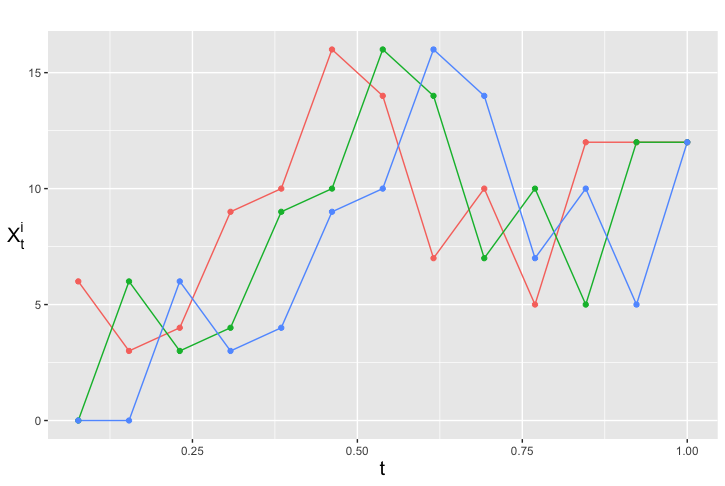}
        \caption{$X^1_t$ (red), $X^3_t$ (green), and $X^5_t$ (blue), coordinates of the lead-lag embedding with lag $2$ against $t$, for $t \in [0,1]$.}
        \label{fig:lead_lag}
    \end{figure}

    Introduced by \citet{chevyrev2016primer} and \citet{flint2016discretely}, the lead-lag transformation has been applied by, e.g., \citet{lyons2014extracting}, \citet{lyons2014feature}, \citet{kormilitzin2016application}, , and \citet{yang2017leveraging}. Building on the time path, the idea is to add lagged versions of the coordinates $X^1$ and $X^2$ as new dimensions. Let $\ell$ be the length of the input path, and $0=t_1< t_2 < \dots < t_\ell<t_{\ell+1}=1$ be a partition of $[0,1]$ into $\ell+1$ points. Then the lead-lag path with lag 1 is defined by
    \begin{align*}
    X: [0,1] &\to \R^5 \\
    t &\mapsto ( X^1_t, X^2_t, t, X^3_t, X^4_t).
    \end{align*}
    In this definition, $X^1$ and $X^2$ are a linear interpolation of the sequence
    \[\big[(x^1_1,x^2_1),\dots, (x^1_\ell,x^2_\ell), (x^1_\ell,x^2_\ell) \big],\]
    in which the last point is repeated twice, and 
    \begin{equation}
    \label{eq:def_lagged_coordinates}
    X^3_t =  \begin{cases} 
    0 & \text{if } t<t_1\\
    X^1_{t-t_1} & \text{otherwise}
    \end{cases}, \qquad X^4_t =  \begin{cases} 
    0 & \text{if } t<t_1\\
    X^2_{t-t_1} & \text{otherwise}
    \end{cases}.
    \end{equation}
    This yields a 5-dimensional path such that the last two coordinates are delayed copies of the first two, with a delay of $t_1$. The process can be iterated, creating a path in $\R^7$ with two lags $t \mapsto ( X^1_t, X^2_t, t, X^3_t, X^4_t, X^5_t, X^6_t)$, where $X^1$ and $X^2$ are linear interpolations of the data with the last point repeated three times, $X^3$ and $X^4$ are defined by \eqref{eq:def_lagged_coordinates}, and 
    \begin{equation*}
    X^5_t =  \begin{cases} 
    0 & \text{if } t<t_2\\
    X^1_{t-t_2} & \text{otherwise}
    \end{cases}, \qquad X^6_t =  \begin{cases} 
    0 & \text{if } t<t_2\\
    X^2_{t-t_2} & \text{otherwise}
    \end{cases} .
    \end{equation*}
    In this way, the lead-lag path can naturally be defined for any lag in $\N^\ast$. This path is highly dimensional (in $\R^7$ for a lag of $2$) and cannot be represented easily. Therefore, Figure \ref{fig:lead_lag} shows some coordinates against time, namely $X^1$, $X^3$, and $X^5$.
    
    \paragraph{Stroke path}
    
    For the Quick, Draw!~data, extra information about pen jumps is provided. In the context of Arabic handwriting recognition, \citet{wilson2018path} have introduced the idea of encoding information about jumps into a new coordinate. In essence, the approach is to use a $3$-dimensional path in which the last dimension corresponds to strokes, in a similar way to the encoding of matrix \eqref{eq:quick_draw_matrix}. This procedure can be deployed in various ways and we restrict our attention to three of them.
    
    The first approach uses the description of a drawing as given in \eqref{eq:quick_draw_matrix}. Recall that, in this matrix, the stroke categorical variable is initialized to $1$, and increased by $1$ each time a different stroke begins. The idea is then to linearly interpolate the columns of the matrix, considered as points in $\R^3$. As can be seen in Figure \ref{fig:flower_stroke_3d_path}, each stroke is then represented in a different horizontal plane. This procedure looks like the most natural way to encode jumps information, and from now on is called ``version 1" of the stroke path. 
    
    A related approach is considered by \citet{wilson2018path}. It is represented in Figure \ref{fig:flower_stroke_3d_path_2} and subsequently called ``version 2". Here, each stroke is indexed by odd integers, that is the first stroke is indexed by $1$, the second by $3$,..., and the $k$th by $2k-1$. Two intermediary points are added between each stroke, indexed by even integers. For example, if $(x^1_{\ell_1}, x^2_{\ell_1},1)$ is the last point of the first stroke, and $(x^1_{\ell_1+1}, x^2_{\ell_1+1},3)$ is the first point of the second stroke, the points $(x^1_{\ell_1}, x^2_{\ell_1},2)$ and $(x^1_{\ell_1+1}, x^2_{\ell_1+1},2)$ are added to the path and linearly interpolated. This is represented in Figure \ref{fig:flower_stroke_3d_path_2}. The only difference with the previous path is how the path moves from one plane to another one. Instead of doing a straight line, it moves in two steps: one in a horizontal plane and another one parallel to the vertical axis. With this embedding and a recurrent network, \citet{wilson2018path} have achieved a significant decrease in the error rate of Arabic characters recognition.
    
    Finally, for comparison purposes, a strictly monotone coordinate is also considered. It has jumps of $1$ when a new stroke begins, and otherwise grows linearly inside one stroke, such that it has increased by $1$ between the beginning and the end of the stroke. In this definition, the goal is to check whether having a strictly monotone coordinate increases accuracy, while in the two previous versions the stroke coordinate is piecewise constant. This embedding can be seen as a mix between time and stroke paths and could inherit the good properties of both. The resulting path is called ``version 3" and shown in Figure \ref{fig:flower_3d_path_time_stroke}. 
    
    To conclude, there exists a broad range of embeddings, living in spaces of various dimensions. They lead to different signature features, which therefore do not have the same statistical properties. The embedding choice will prove to have a significant influence on accuracy.
    
    \subsection{Results}
    \label{sec:path_embedding_study}
    
    In this subsection, the results of our study on embedding performance are presented. To this end, the first approach described in Section \ref{sec:signature_ml_review} is implemented. Starting from the raw data, it is first embedded into a continuous path, then its truncated signature is computed and used as input for a learning algorithm. The embeddings described in the previous section are used. Note that the lead-lag path is taken with lag $1$, but other lags will be discussed in Section \ref{sec:performance_signature}. Each feature is normalized by the absolute value of its maximum so that all input values lie in $[-1,1]$. The findings should be independent of the data and the underlying statistical model so a range of different algorithms is used. Their hyperparameters have been set to their default values, without trying to optimize them for each dataset. Indeed, the goal is not to select the best algorithm or to achieve a particularly good accuracy, but rather to compare the performance of different embeddings. The classification metric to assess prediction quality is the accuracy score. Denoting by $(y_1, \dots , y_{n_{\text{test}}})$ the test set's labels, and $(\hat{y}_1, \dots, \hat{y}_{n_{\text{test}}})$ the predicted labels, this score is defined by
    \begin{equation}
    \label{eq:def_accuracy}
    \text{Acc}_{\text{test}} = \frac{1}{n_{\text{test}}} \sum_{i=1}^{n_{\text{test}}} {\mathbb{1}}_{\hat{y}_i=y_i} .
    \end{equation}
    The four following algorithms have been used throughout the study.
    
    \begin{itemize}
        \item Following \citet{yang2017leveraging}, a dense network with one hidden layer composed of 64 units with linear activation functions is first considered. A softmax output layer and the categorical cross-entropy loss are used, which yields a linear model equivalent to logistic regression. This architecture is a sensible choice, since Proposition \ref{prop:linear_approx} states that linear functions of the signature approximate arbitrarily well any continuous function of the input path. The Python library \pkg{keras} \citep{chollet2015keras}, with TensorFlow backend, is used. The network is regularized by adding a dropout layer after the input layer, with a rate of 0.5. Optimization is done with stochastic gradient descent with an initial learning rate of 1. It is reduced by $2$ when no improvement is seen on a validation set during 10 consecutive epochs. The maximal number of epochs is set to 200 and the mini-batch size to 128.
        
        \item Furthermore, the performance of a random forest classifier with 50 trees, implemented in scikit-learn \citep{scikit-learn}, is tested. It is a nonlinear very popular method initially proposed by \citet{Breiman2001}.
        
        \item The XGBoost algorithm, introduced by \citet{chen2016xgboost}, and implemented in the Python package \pkg{xgboost}, is also used. It is a state-of-the-art gradient boosting technique, building upon the work of \cite{friedman2001greedy}. The maximum number of iterations is set to 100 and early stopping with a patience of $5$ is used to prevent overfitting and speed up training. The maximum depth of a tree is set to 3 and the minimum loss reduction to make a split to 0.5.
        
        \item Finally, a nearest neighbor classifier is run with a default value of 5 neighbors. This method is known to suffer from the curse of dimensionality, so it is of interest to see how the signature truncation order affects its performance.
    \end{itemize}
    For each of the algorithms described above and each dataset of Section \ref{sec:datasets} (Quick, Draw!, Urban Sound, and Motion Sense), the following steps are repeated:
    \begin{enumerate}
        \item Split the data into training, validation, and test sets, as described in Table \ref{tab:dataset_summary}.
        \item Choose an embedding and transform samples $\mathbf{x_i}$ into continuous paths $X_i:[0,1] \to \R^d$.
        \item For $k=1,\dots,K$:
        \begin{enumerate}
            \item Compute $S_k(X_i)$, the signature truncated at order $k$, for every sample $i$. This results in training, validation and test sets of the form
            \[\big\{S_k(X_1), \dots, S_k(X_n)\big\}, \]
            where  $S_k(X_i) \in \R^{\frac{d^{k+1}-1}{d-1}}$  if $d>1$, and $\R^{k}$ if $d=1$.
            \item Fit the algorithm on the training data. Validation data is used when the algorithm chosen is the linear neural network or XGboost, to adapt the learning rate and to implement early stopping, respectively.
            \item Compute the accuracy, defined by \eqref{eq:def_accuracy}, on the test set.
        \end{enumerate}
    \end{enumerate}
    The maximal value considered for the truncation order, denoted by $K$, is fixed so that the number of features are computationally reasonable. The meaning of ``reasonable'' depends on each dataset, as they have a different number of samples and classes. For Quick, Draw!, we will consider up to $10^5$ samples, for Urban Sound $10^4$ and for Motion Sense $5 \times 10^5$. 

    For a path $X$ in $\R^d$, the number of features is equal to $(d^{k+1}-1)/(d-1)$ if $d>1$, and to $k$ if $d=1$ (see Table \ref{tab:truncation_dimensions} for some values). Therefore, the number of features depends on the dimension $d$ of the embedding and the truncation order $k$. But $d$ is different depending on the dataset and the embedding. Thus, to compare the quality of different embeddings, the accuracy score is plotted against the log number of features, which yields one curve per embedding, where each point corresponds to a different truncation order $k$. One embedding curve being above the others means that, at equal input size, this embedding performs better.

    \begin{figure}[h]
        \centering
        \includegraphics[width=\textwidth]{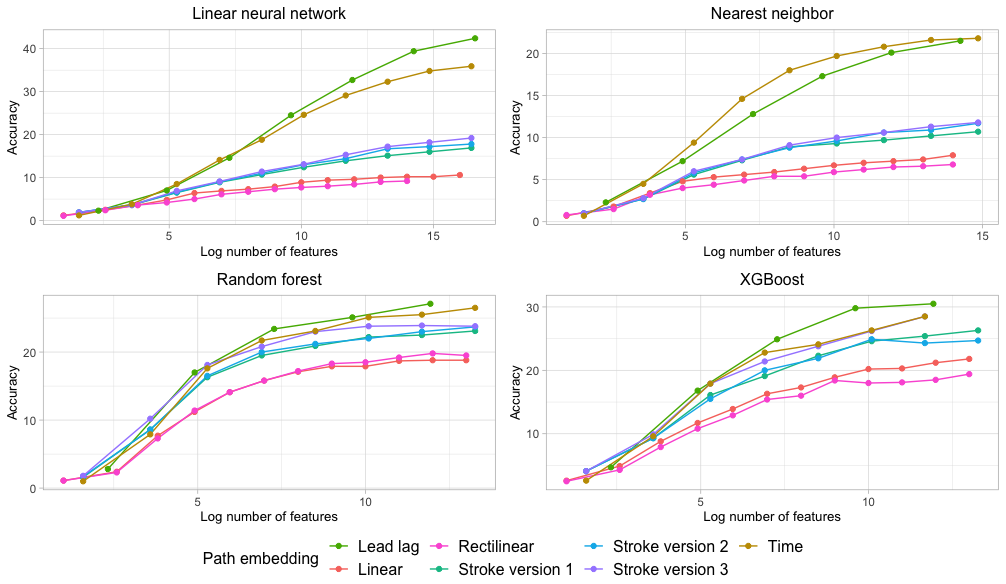}
        \caption{Quick, Draw!~dataset: prediction accuracy on the test set, for different algorithms and embeddings.}
        \label{fig:quick_draw_embedding_results}
    \end{figure}
    \begin{figure}[h!]
        \centering
        \includegraphics[width=\textwidth]{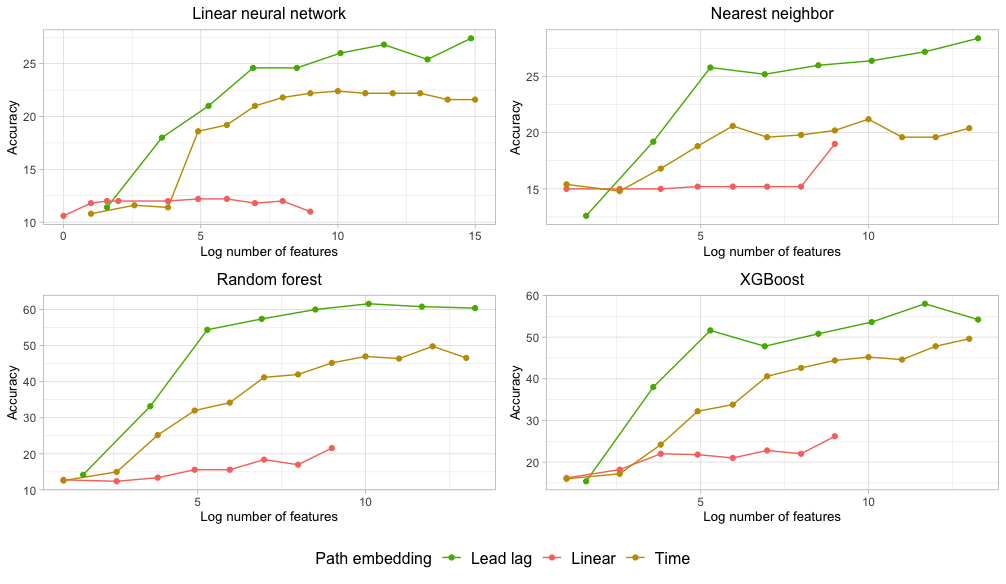}
        \caption{Urban Sound dataset: prediction accuracy on the test set, for different algorithms and embeddings.}
        \label{fig:urban_sound_embedding_results}
    \end{figure}
    \begin{figure}[h!]
        \centering
        \includegraphics[width=\textwidth]{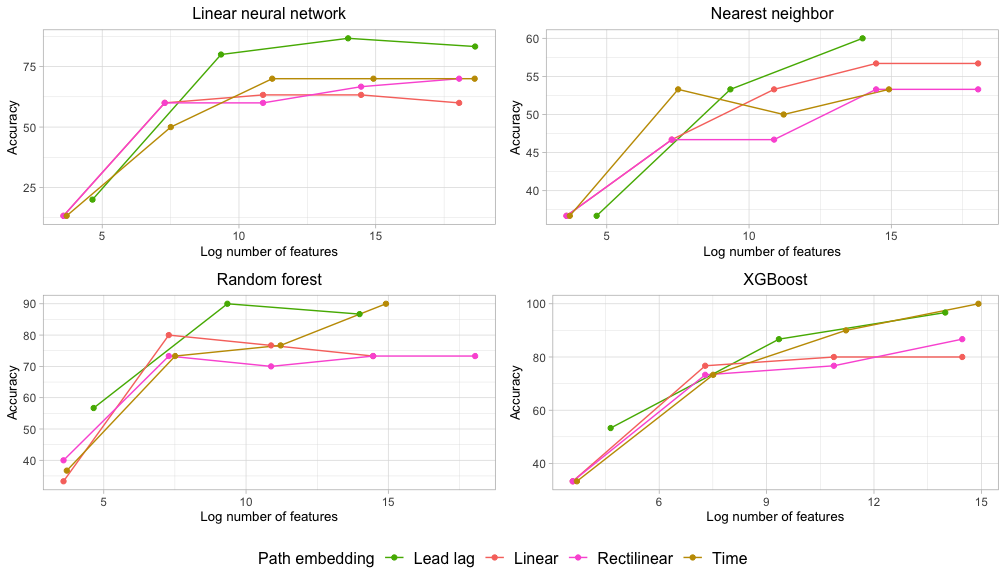}
        \caption{Motion Sense dataset: prediction accuracy on the test set, for different algorithms and embeddings.}
        \label{fig:motion_sense_embedding_results}
    \end{figure}
    
    The results of this procedure are plotted in Figures \ref{fig:quick_draw_embedding_results}, \ref{fig:urban_sound_embedding_results} and \ref{fig:motion_sense_embedding_results}, which correspond respectively to the Quick, Draw!, Urban Sound and Motion Sense datasets. A first observation is that some embeddings, namely the time and lead-lag, seem consistently better, whatever the algorithm and the data used. It suggests that this performance is due to the intrinsic theoretical properties of signatures and embeddings, not to domain-specific characteristics. It is particularly remarkable as the dimension of input streams is different from one dataset to another. 
    
    The linear and rectilinear embeddings (red and pink curves), which are often used in the literature, appear to give the worst results. These two interpolation methods do not differ much in their results, although the linear path seems to be slightly better. Moreover, it seems that the smaller the dimension $d$, the worse their performance. Indeed, the linear embedding is especially bad for the Urban Sound dataset, which is unidimensional, whilst the difference is less pronounced for the Motion Sense dataset, which has values in $\R^{12}$. This bad performance can be explained by the fact that there is no guarantee that the signature transformation is unique when using the linear or rectilinear embeddings. Therefore, two different paths can have the same signature, without necessarily corresponding to the same class. 
    
    On the other hand, the best embedding is the lead-lag path (green curve), followed closely by the time path (brown curve). The difference between these two embeddings is again most important for the Urban Sound dataset. For the Quick, Draw!~data, stroke paths have intermediate results, better than the linear path but still worse than the time and lead-lag paths. Yet stroke paths are the only embeddings in which new information, about pen jumps, is included. It is surprising how little impact this information seems to have on prediction accuracy. Note that in all of these cases, the uniqueness of the signature is ensured so it cannot explain the performance differences. 
    
    Good performance of the lead-lag path has already been noticed in the literature. However, up to our knowledge, there are few theoretical results. Still, \citet{flint2016discretely} have considered a discretely sampled input path $X$, assumed to be a continuous semimartingale, and have studied convergence results of its associated lead-lag path, called Hoff process, when sampling frequency increases. Thus, a lot of questions remain open concerning the statistical performance of the time and lead-lag embeddings, with, to our knowledge, no theoretical result in classification or regression frameworks.
    
    To conclude this section, the take-home message is that using the lead-lag embedding seems to be the best choice, regardless of the data and algorithm used. It does not cost much computationally and can drastically improve prediction accuracy. Moreover, the linear and stroke paths yield surprisingly poor results, despite their frequent use in the literature.

    \subsection{Running times}

    \begin{figure}[h]
        \centering
        \includegraphics[width=\textwidth]{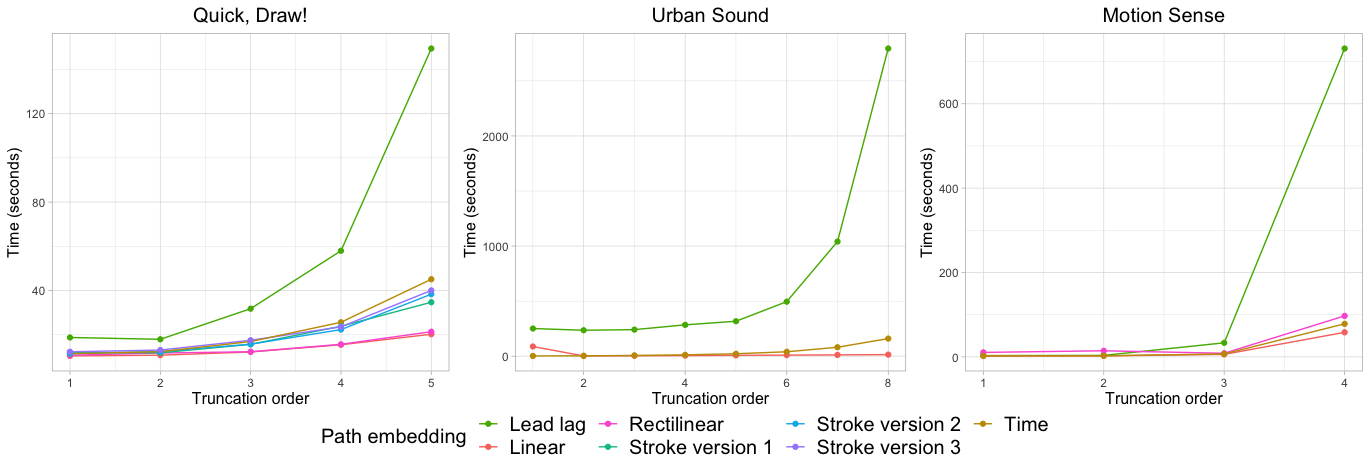}
        \caption{Running time (in seconds) to compute signature features and fit a random forest classifier for various embeddings, as a function of the truncation order. }
        \label{fig:running_times_rf}
    \end{figure}

    To conclude this study on embeddings, this section presents some results on the computational complexity of the different embeddings and truncation order. In Figure \ref{fig:running_times_rf}, the running times for computing signature features and fitting a random forest are shown as functions of the truncation order for various embeddings. The experiments were run on 32 Intel Xeon E5-4660 cores and parallelized with the \texttt{multiprocessing} Python package. The most expensive embedding is the lead-lag, which is not surprising as it doubles the path dimension. Moreover, increasing the truncation order increases exponentially the running time. For Quick, Draw! the running time is of the order of 10-100 seconds, for Motion Sense of the order of 100 seconds and for Urban Sound around 1000 seconds. This is directly linked to the length of the series: the longer the series, the more expensive it is to compute signatures.

    \begin{figure}[h]
        \centering
        \includegraphics[width=\textwidth]{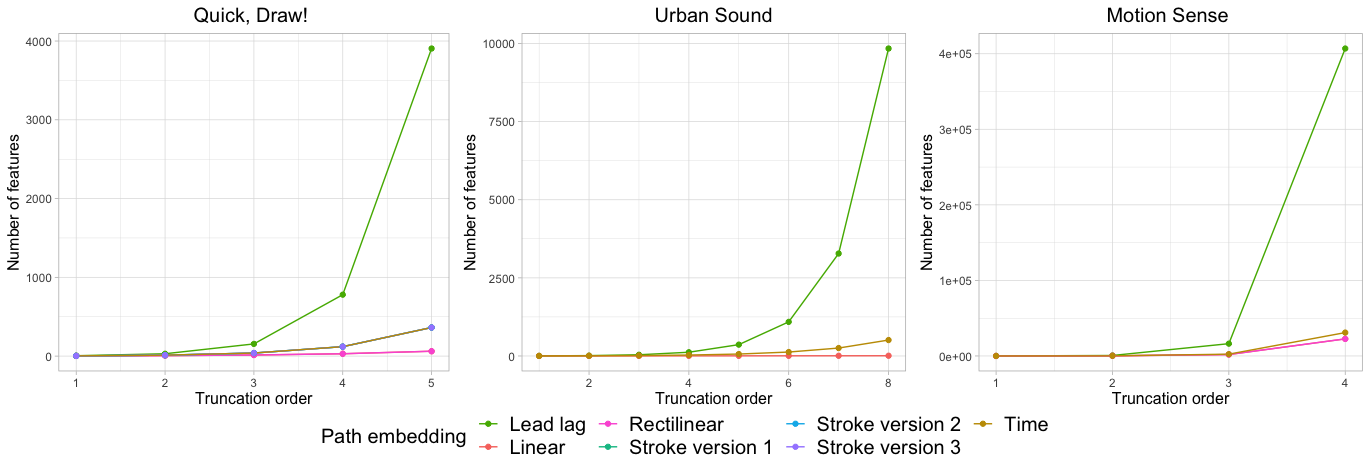}
        \caption{Number of features of various embeddings as a function of the truncation order. }
        \label{fig:memory_rf}
    \end{figure}

    In Figure \ref{fig:memory_rf} is presented the number of input features for each combination of embedding and truncation order. This is proportional to the memory needed to run each experiment. As given by equation \eqref{eq:size_signature}, it is clear that the storage cost increases exponentially with the truncation order, which is the main limitation of the signature method.
    
    \section{Simulation study of autoregressive processes}
    \label{sec:simu_ar}

\begin{figure}[h]
        \centering
        \begin{subfigure}[b]{0.32\textwidth}
            \includegraphics[width=\textwidth]{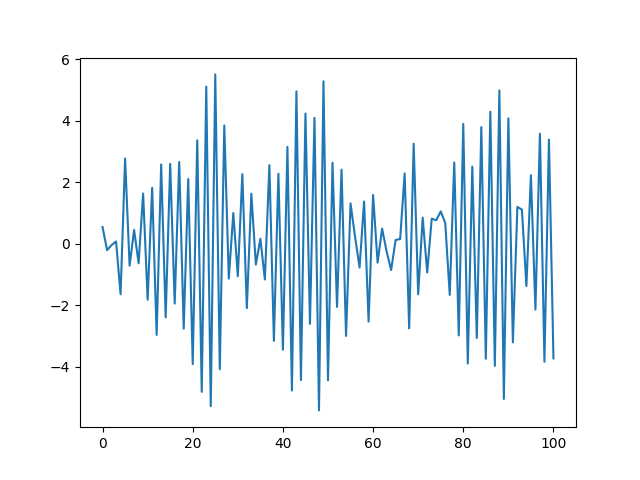}
            \caption{AR(1): $\phi_1=-0.9$}
            \label{fig:ar_1_example}
        \end{subfigure}%
        ~
        \begin{subfigure}[b]{0.32\textwidth}
            \centering
            \includegraphics[width=\textwidth]{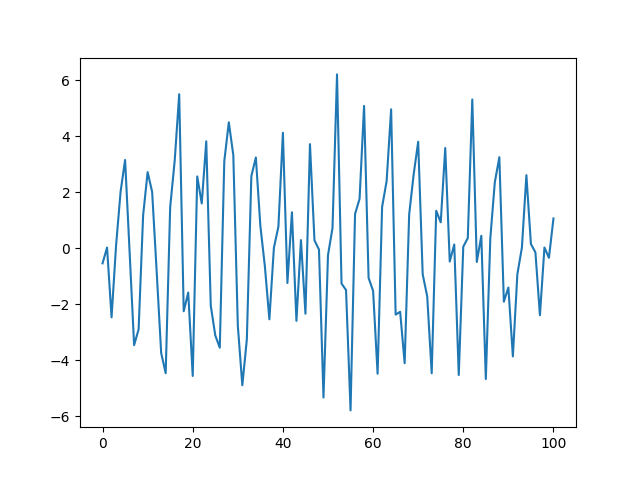}
            \caption{AR(3): $\phi_1=\phi_2=0, \phi_3=-0.9$}
            \label{fig:ar_3_example}
        \end{subfigure}
        ~
        \begin{subfigure}[b]{0.32\textwidth}
            \centering
            \includegraphics[width=\textwidth]{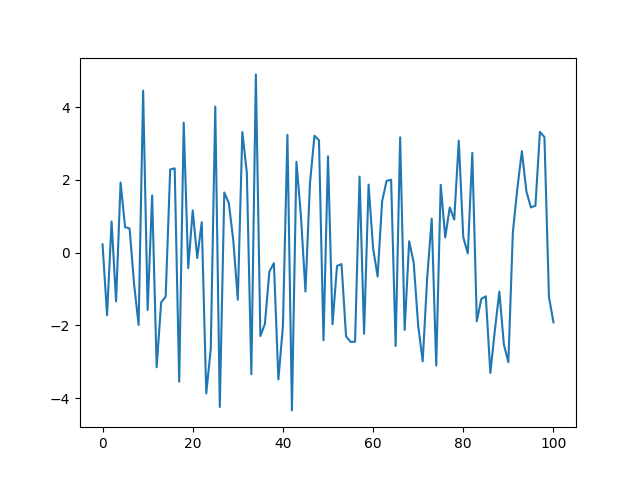}
            \caption{AR(8): $\phi_1=\dots=\phi_7=0, \phi_8=-0.9$}
            \label{fig:ar_8_example}
        \end{subfigure}
        \caption{Sample paths of an AR(p) process.}
        \label{fig:ar_examples}
    \end{figure}

    In order to confirm the previous findings on the performance of the lead-lag embedding, we undertake a simulation study with autoregressive processes. This study will also provide insights on the sensitivity of the method to some hyperparameters such as the lag and the signature truncation order. We place ourselves in a regression setting, that is we consider $n$ realizations of an AR($p$) process $Z_t$, observed on $\ell=100$ time points, and defined by
    \begin{equation}
    \label{eq:def_ar_p}
    Z_t= \phi_1 Z_{t-1} + \dots + \phi_p Z_{t-p} + \varepsilon_t, \quad 1 \leq t \leq \ell,
    \end{equation}
    where $\varepsilon_t$ is a gaussian random variable with mean 0 and variance 1. Some examples are shown in Figure \ref{fig:ar_examples}. Our goal is to predict the next time step, that is $y=Z_{\ell +1}$, with a linear regression and signature features. The input data is therefore a set $\{(\mathbf{x_1},y_1), \dots, (\mathbf{x_n}, y_n) \}$, where each $\mathbf{x_i}$ is one realization of an AR(p) process:
    \[\mathbf{x_i}= \begin{pmatrix}Z_{i,1}, \dots, Z_{i,\ell}  \end{pmatrix} \in \R^{\ell \times 1},\]
    and $y_i=Z_{i,\ell+1}$, where $Z_{i,\cdot}$ follows \eqref{eq:def_ar_p}.

    \begin{figure}[h]
        \centering
        \includegraphics[width=\textwidth]{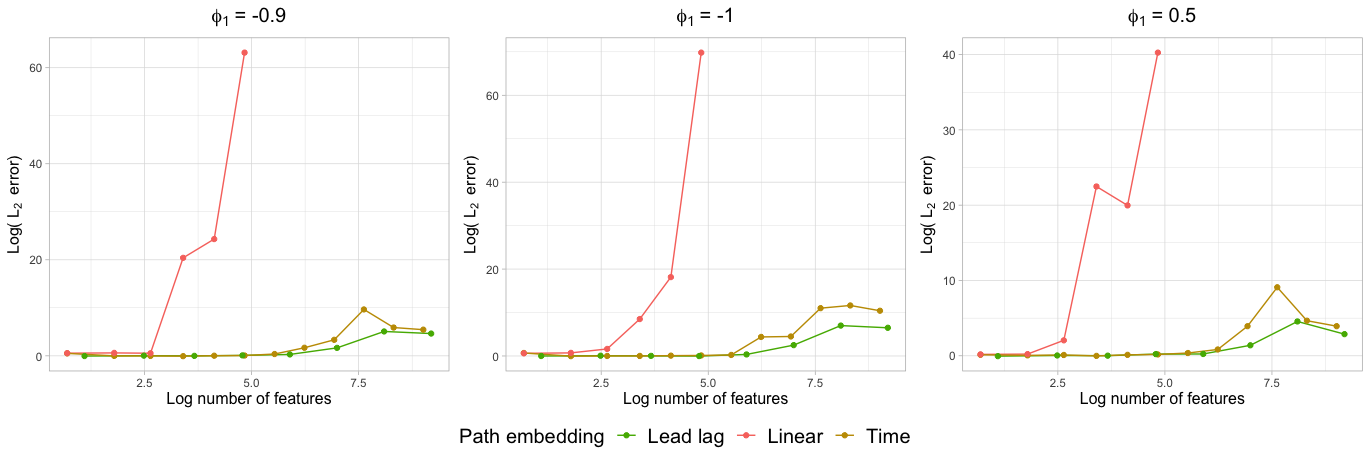}
        \caption{Logarithm of the test error for different embeddings and truncation orders for prediction of an AR(1) process. The left panel corresponds to $\phi_1=-0.9$, the middle one to $\phi_1=-1$ and the right one to $\phi_1=0.5$.}
        \label{fig:ar_1_embedding_results}
    \end{figure}

    First, Figure \ref{fig:ar_1_embedding_results} shows the results of the same study on the embeddings performance as in Section \ref{sec:embedding} for different AR(1) processes. The parameter $\phi_1$ is equal to $-0.9$, $-1$ and $0.5$, in order to obtain both stationary and nonstationary models. The performance of the different embeddings is plotted against a range of truncation orders. The metric is the $L_2$ error on a test set, which is defined by 
    \[S=\frac{1}{n_{\textnormal{test}}} \sum_{i=1}^{n_{\textnormal{test}}} (y_i - \hat{y}_i)^2,\]
    with the same notations as \eqref{eq:def_accuracy}. Contrary to Figures \ref{fig:quick_draw_embedding_results}, \ref{fig:urban_sound_embedding_results} and \ref{fig:motion_sense_embedding_results} which use as metric the accuracy, the smaller $S$ the better the prediction. It is clear in Figure \ref{fig:ar_1_embedding_results} that the time and lead-lag embeddings have the smallest errors, confirming the findings of the previous section on real-world datasets. Moreover, the figures are very similar for stationary or nonstationary series, and different strength of time dependence. This shows the generality of the signature method, which does not require strong assumptions on the law of the underlying process.

    \begin{figure}[h]
        \centering
        \includegraphics[width=\textwidth]{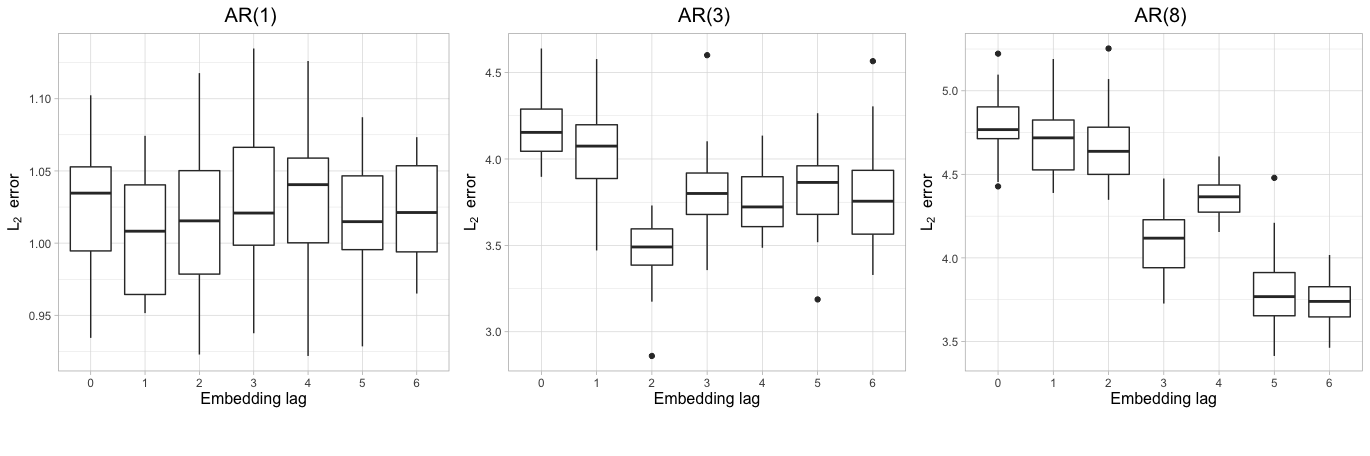}
        \caption{Error boxplots of the $L_2$ error as a function of the lag of the embedding, for different AR processes.}
        \label{fig:ar_lag_boxplots}
    \end{figure}
    
    A natural question is whether the lag parameter is linked to the dependence in the time series. To tackle this issue, Figure \ref{fig:ar_lag_boxplots} shows a boxplot of the $L_2$ error as a function of the lag for 3 different values of $p$. The parameters of model \eqref{eq:def_ar_p} are set to the following values: for $p=1$, $\phi_1=-0.9$; for $p=3$, $\phi_1=\phi_2=0$ and $\phi_3=-0.9$; for $p=8$, $\phi_1=\dots=\phi_7=0$ and $\phi_8=-0.9$.  For each lag, the best truncation order is selected with a validation set, that is the truncation order is chosen to be the one achieving the lowest error on a validation set. The procedure is then evaluated on another test set. This procedure is iterated 20 times to obtain estimates of the variability of the error. Figure \ref{fig:ar_lag_boxplots} shows that when $p$ increases, the best lag increases. For $p=1$, all lags seem to achieve similar errors, for $p=3$, there is an error jump between a lag of 1 and a lag of 2, and for $p=8$ the error decreases towards the best lag of 6. This is strong evidence of the link between the time dependence and the lag parameter.

    \begin{figure}[h]
        \centering
            \includegraphics[width=.5\textwidth]{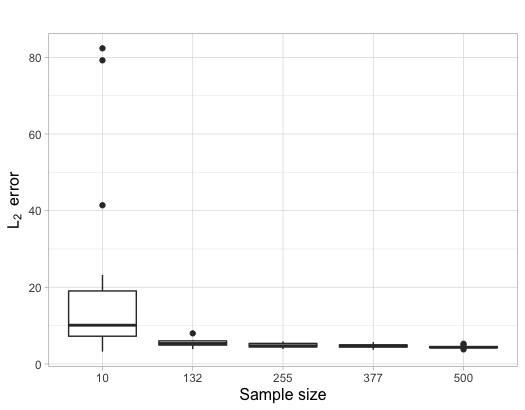}
            \caption{Error boxplot for different sample sizes.}
            \label{fig:ar_3_error}
    \end{figure}

\begin{figure}[h]
        \centering
        \includegraphics[width=.8\textwidth]{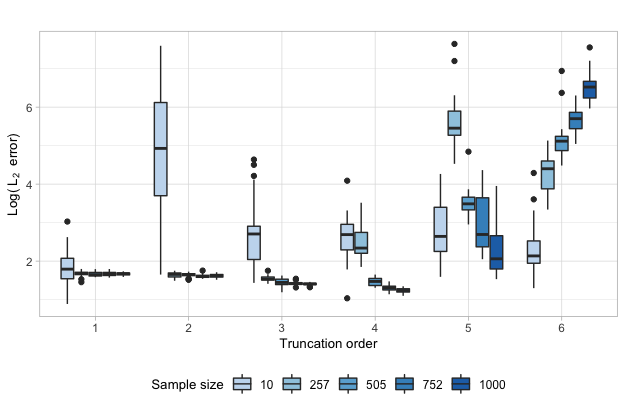}
        \caption{Average error for different truncation orders and sample sizes.}
        \label{fig:ar_cv_curves_sample_size}
    \end{figure}

    Finally, Figures \ref{fig:ar_3_error} and \ref{fig:ar_cv_curves_sample_size} investigate the link between the prediction error, the sample size and the truncation order of the signature for an AR(3) process with the same parameters as before. We choose a lead-lag embedding with a lag of 2, as suggested by Figure \ref{fig:ar_lag_boxplots}. For each sample size and truncation order, the model is fitted and evaluated 20 times. In Figure \ref{fig:ar_3_error}, the truncation order is chosen as the minimizer of the error on a validation set. Then, a boxplot of the errors is plotted as a function of the sample size. Both the error and its variance decrease fast when the sample size increases. 

    On the other hand, Figure \ref{fig:ar_cv_curves_sample_size} shows error boxplots as a function of the truncation order $K$, for various sample sizes. When the sample size is large enough, typically larger than 200, a bias-variance tradeoff can be observed: the error first decreases with the truncation order until a minimum is reached and then the error and its variance increase fast because the number of covariates is too large compared to the number of observations. It is interesting to see that the best truncation order increases when the sample size increases but stabilizes at $K=4$. Moreover, the error variance is very large for a sample size of 10 but stays reasonably small for larger sample sizes.

    \section{Signature domain and performance}
    \label{sec:sig_domain_and_performance}

    \subsection{Comparison of local and global signature features}
    \label{sec:subpaths_signature}
    
    As discussed in Section \ref{sec:signature_ml_review}, several authors do not compute signatures on the whole time interval but use instead a partition of $[0,1]$. The rationale for this division is to describe the path by a sequence of truncated signatures, rather than one signature computed over the whole domain. Therefore, it is not surprising that this approach is typically used in combination with recurrent neural networks. In this section, it is investigated whether signatures computed on a sub-interval contain some local information not present in signatures of the whole path. To this end, following \citet{wilson2018path}, a dyadic partition of $[0,1]$ is considered, and defined by \eqref{eq:dyadic_partition}:
    \[0 \leq 2^{-q} < \dots < j2^{-q} < \dots < (2^q-1) 2^{-q} \leq 1, \quad 0 < j \leq 2^q \]
    where $q$ is the dyadic order. For different values of $q$, signature coefficients are computed on each interval $\big[(j-1) 2^{-q}, j2^{-q} \big]$ of the dyadic partition. Therefore, for each input path $X_i$, a collection of signature vectors is obtained, which is then stacked into one large vector. This vector is then the input of a learning algorithm, and the prediction accuracy curves of different dyadic orders are compared. The time embedding with the linear neural network described in Section \ref{sec:path_embedding_study} is used. This process is summarized below.
    \begin{enumerate}
        \item Split the data into training, validation and test sets.
        \item For $q=0,\dots, Q$, and $k=1,\dots,K$:
        \begin{enumerate}
            \item[$(i)$] For $j=1,\dots, 2^q$, compute the signature truncated at order $k$ on $\big[(j-1)2^{-q}, j2^{-q}\big]$, denoted by  \[S_k(X_i)_{[(j-1)2^{-q}, j2^{-q}]},\] where $X_i$ is the time embedding of  sample $\mathbf{x_i}$. Repeat this over all training samples.
            \item[$(ii)$]  For each training sample $X_i$, concatenate all signature vectors and obtain one vector $\tilde{S}_k(X_i)$ containing all $S_k(X_i)_{[(j-1)2^{-q}, j2^{-q}]}$, for $j=1,\dots ,2^q$. This yields a dataset \[\big\{\tilde{S}_k(X_1), \dots, \tilde{S}_k(X_n)\big\}. \]
            \item[$(iii)$]  Fit a linear neural network with this data as features.
            \item[$(iv)$]  Compute accuracy on the test set.
        \end{enumerate}
    \end{enumerate}
    \begin{figure}[h]
        \centering
        \includegraphics[width=\textwidth]{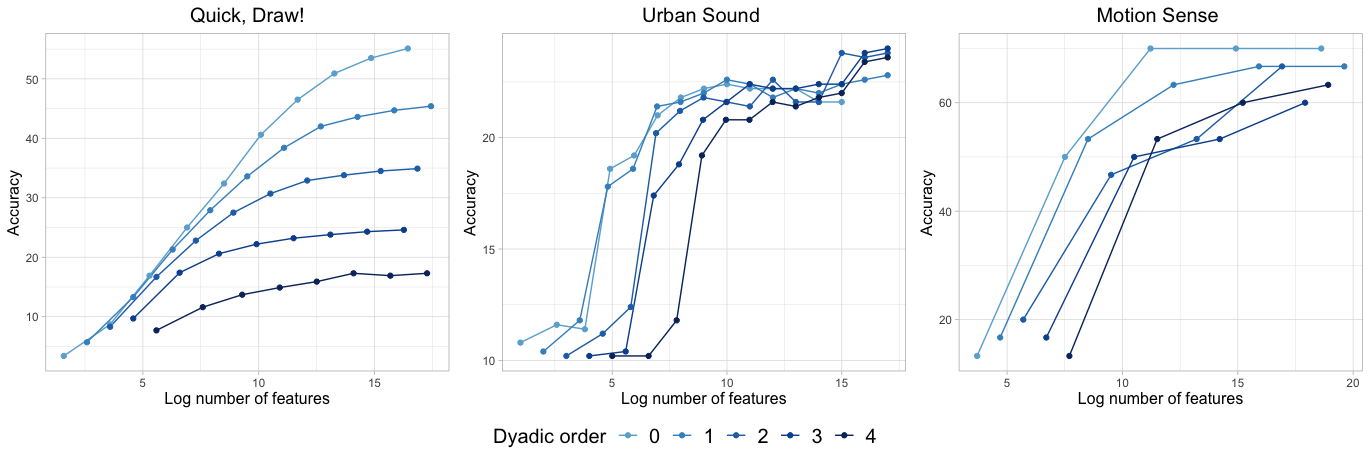}
        \caption{Test accuracy for different dyadic partitions of the path.}
        \label{fig:dyadic_partition_results}
    \end{figure}
    The results of this procedure are shown in Figure \ref{fig:dyadic_partition_results}. First, it is clear that a dyadic order of $0$, which corresponds to computing the signature on the whole interval $[0,1]$, yields the best results. Indeed, the curve is always above the others for the Quick, Draw!~and Motion Sense datasets. This is less obvious for the Urban Sound dataset, as the curve $q=0$ is really close to the one corresponding to a dyadic order of $1$. However, it still achieves a better result for most truncation orders $k$. This difference between datasets can be linked to their length: it seems that the longer the series, the better the accuracy of thin dyadic partitions. Indeed, high dyadic orders perform best for the Urban Sound dataset, which has an average of 170 000 sampled time points (see Table \ref{tab:dataset_summary}), whereas it is clear that each new dyadic split decreases accuracy for the Quick, Draw!~data, which has an average of 44 sampled points.  
    
    In a nutshell, little local information seems to be lost when the signature of the whole path is computed. However, it may be worth considering partitions of the path for long streams. 
    
    \subsection{Performance of the signature}
    \label{sec:performance_signature}
    
    The message of previous sections is that the lead-lag embedding is the most appropriate in a learning context and that signatures should be computed over the whole path domain. As a natural continuation, the effect of the algorithm is now examined more closely and the prediction scores are compared to the literature. It turns out that the signature combined with a lead-lag embedding has an excellent representation power, to the extent that it achieves prediction scores close to state-of-the-art methods, without using any domain-specific knowledge.
    
\begin{figure}[h]
        \centering
        \includegraphics[width=\textwidth]{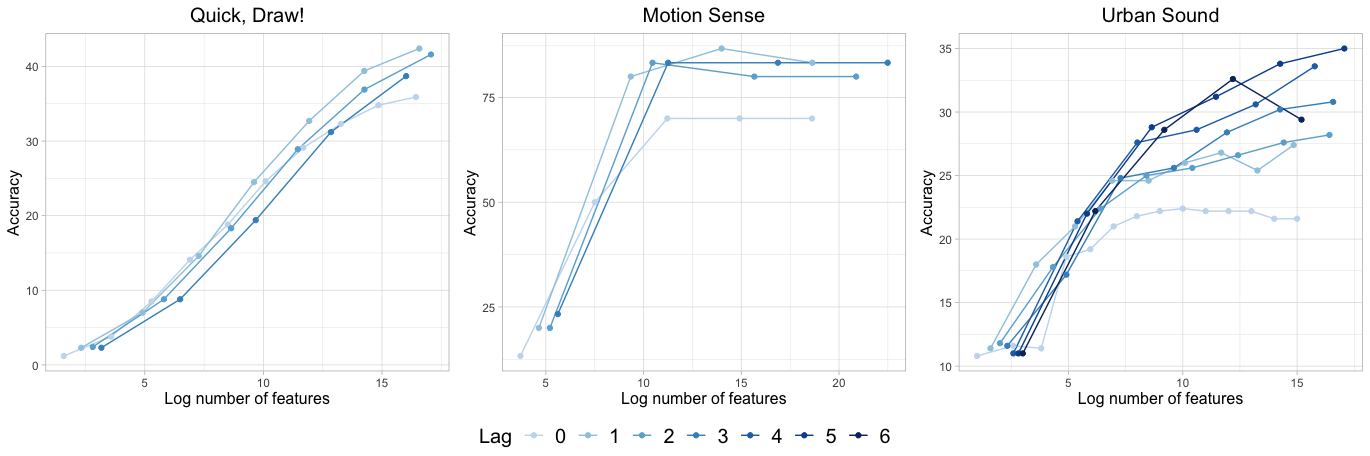}
        \caption{Test accuracy for different lags.}
        \label{fig:lag_selection}
    \end{figure}

    Before starting the comparison, it is worth pointing out that the lead-lag embedding has a hyperparameter that has not yet be tuned, which is the number of lags. It is now selected with the same approach as in previous sections: for each lag, the test accuracy is plotted against the number of features for various truncation orders. The lag which gives a curve above the others is selected. Figure \ref{fig:lag_selection} highlights that curves overlap for the Motion Sense and Urban sound cases, therefore, when there is a doubt on which curve is above, the smallest lag is picked. For the Quick, Draw!~and Motion Sense datasets, the best lag is then 1, whereas it is 5 for the Urban Sound dataset. 
    
    Finally, the truncation order is selected with a validation set: the truncation order achieving the highest accuracy on a validation set is picked. In general, the truncation order and the number of lags could also be selected with cross-validation.
    
    The Motion Sense and Urban Sound datasets do not require a lot of computational resources, as they have a reasonable size (a few hundred samples for Motion Sense and thousand for Urban Sound---see Table \ref{tab:dataset_summary}) and a small number of classes. On the other hand, the Quick, Draw!~recognition task, which is comprised of 340 classes, is more involved and requires an elaborate algorithm as well as a significant number of training samples. Therefore, more data will be used than in the previous sections: $12\, 185\, 600$ training samples and $87\, 040$ validation samples.
    
    For the Quick, Draw!~dataset, our results are compared to a Kaggle competition \citep{quickdrawkaggle}. In this competition, 1 316 teams competed for a prize of 25 000\textdollar. State-of-the-art deep convolutional networks, such as MobileNet or ResNet, trained with several millions of samples, were among the best competitors. Teams on the podium used ensembles of such networks. These winning methods require a lot of computing resources and are specific to images. The metric used in the competition was the mean average precision, defined as follows. Denoting by  $\{y_1, \dots ,y_{n_{\text{test}}}\}$ the test set labels, three ranked predictions are made for each sample, denoted by 
    \[\big\{(\hat{y}^1_{1},\hat{y}^2_{1},\hat{y}^3_{1}), \dots , (\hat{y}^1_{n_{\text{test}}},\hat{y}^2_{n_{\text{test}}},\hat{y}^3_{n_{\text{test}}}) \big\},\] 
    where $\hat{y}^1_i$ is the class with the largest probability, $\hat{y}^2_i$ the second largest, and so on. Then, the mean average precision at rank 3 is defined by
    \[MAP_3= \frac{1}{n_{\text{test}}} \sum_{i=1}^{n_{\text{test}}} \sum_{j=1}^{3} \frac{\mathbb{1}\{\hat{y}^j_i=y_i  \} }{j}.\]
    Mean average precision is computed by the competition platform on 91\% of a test set of 112 200 samples. The small neural network used in Section \ref{sec:path_embedding_study} is enhanced by using ReLU activation functions and adding three hidden layers with 256 nodes. The network is trained during 300 epochs with an Adam optimizer. At each epoch, it is trained on 609 280 samples randomly selected among the 12 185 600 training samples. The best team obtains a $MAP_3$ of 95\% whereas this small network combined with signature features truncated at order 6 already achieves 54\%. The winners use an ensemble of several dozens of deep neural networks, trained on 49 million samples. This kind of architectures requires considerably more computational capacities than ours.
    
    For the Urban Sound dataset, state-of-the-art results are obtained by \citet{ye2017urban}. The authors combine feature extraction with a mixture of expert models and achieve 77.36 \% accuracy, defined by \eqref{eq:def_accuracy}. The feature extraction step is specific to sound data and is based on several ingredients, such as whitened spectrogram, dictionary learning, soft-thresholding, recurrence quantification analysis, and so on. These crafting operations make use of a lot of domain-specific knowledge and cannot be extended easily to other applications. On the other hand, it is clear from Figure \ref{fig:urban_sound_embedding_results} that a random forest classifier performs well with signature features. Therefore, its hyperparameters are tuned with a lead-lag embedding, a lag of 5, and a signature truncated at order 5. An accuracy of 70 \% is obtained with 460 trees with a maximum depth of 30 and in which 500 random features are considered at each split. 
    
    Finally, \citet{malekzadeh2019mobile} tackle the problem of mobile sensor data anonymization. They build a deep neural network architecture that preserves user privacy but still detects the activity performed. The architecture is built on autoencoders combined with a multi-objective loss function. There is a trade-off between activity recognition and privacy but good activity recognition results are achieved. Performance of the classifier of \citet{malekzadeh2019mobile} is measured with the average $F_1$ score, defined as follows. Assume there are $C$ different classes, and denote by $(y_1, \dots ,y_{n_{\text{test}}})$ the test labels, and by $(\hat{y}_1, \dots , \hat{y}_{n_{\text{test}}})$ the predicted ones. Then, the $F_1$ score is defined by
    \[F_1 = \frac{1}{C} \sum_{c=1}^{C} \frac{ 2 \cdot \text{Precision}_{c} \cdot \text{Recall}_{c}}{\text{Precision}_{c} + \text{Recall}_{c}}, \]
    where 
    \[ \text{Precision}_{c} = \frac{\sum_{i=1}^{n} \mathbb{1}\{\hat{y}_i=y_i=c \} }{\sum_{i=1}^{n} \mathbb{1}\{\hat{y}_i=c \} } \text{ and } \text{Recall}_{c} = \frac{\sum_{i=1}^{n} \mathbb{1}\{\hat{y}_i=y_i=c \} }{\sum_{i=1}^{n} \mathbb{1}\{y_i=c \} }.\]
    \citet{malekzadeh2019mobile} report an average $F_1$ score above 92\%, while the signature truncated at order 3 and combined with a XGBoost classifier achieves a $F_1$ score of 93.5\%. These two scores are close, but the signature approach is computationally much less demanding. 
    
    Despite not being tuned to a specific application, the combination signature + generic algorithm achieves results close to the state-of-the-art in several domains, while requiring few computing resources and no domain-specific knowledge. Indeed, it takes approximately 52 seconds to compute the signature at order 3 of 68 000 Quick, Draw!~samples on one core of a laptop, which results in $0.0008$ second per sample. Besides, signature computations can be parallelized, making the approach scalable to big datasets. Lastly, the signature method achieves its best results for the high dimensional Motion Sense dataset, which suggests that it is especially relevant for multidimensional streams.

    \section{Conclusion}
    \label{sec:conclusion}
    
    The signature method is a generic way of creating a feature set for sequential data and has recently caught the machine learning community's attention. Indeed, it yields results competitive with state-of-the-art methods, while being generic, computationally efficient, and able to handle multidimensional series. One of its appealing properties is that it captures geometric properties of the process underlying the data and does not depend on a specific basis. In this paper, its use in a learning context, and several of its successful applications have been reviewed. The use of signatures relies on representing discretely sampled data as continuous paths, a mechanism called embedding. In the literature, authors use various embeddings, without any systematic comparison. We have compared different common embeddings and concluded that the lead-lag seems to be systematically better, whatever the algorithm or dataset used. Moreover, we have pointed out that the signature of the whole path appears to contain as much information as the signature of subpaths, therefore encoding both global and local properties of the input stream.
    
    Our study is a first step towards understanding how signature features can be used in statistics, and a lot of issues remain open, both practical and theoretical. First, it would be of great interest to understand the theoretical statistical properties of embeddings, in particular, to explain the good performance of the lead-lag path. Moreover, in Section \ref{sec:signature_ml_review}, we have seen that signature features may be combined with feedforward, recurrent, or convolutional neural networks. For each of these architectures, the point of view on signature features is different: they are considered respectively as a vector, a temporal process, or an image. A more detailed understanding of these representations would be valuable. Finally, it could be worth investigating the robustness of the signature method when the truncation order becomes large. Indeed, Figures \ref{fig:quick_draw_embedding_results}, \ref{fig:motion_sense_embedding_results}, and \ref{fig:urban_sound_embedding_results} suggest that the signature may be robust to dimension: the accuracy curves do not decrease when the number of features becomes large, even when a nearest neighbor algorithm is used with more than a hundred thousand features. This phenomenon may deserve a more in-depth study.

\section*{Acknowledgements}

This work was supported by a grant from Région Ile-de-France. I thank Gérard Biau (Sorbonne Université), Benoît Cadre (Université Rennes 2) and Terry Lyons (Oxford University) for stimulating discussions and insightful suggestions; and Patrick Kidger (Oxford University) for his thorough proofreading of the manuscript. I also thank the Editor, the Associate Editor, and two anonymous referees for their careful reading of the paper and constructive comments, which led to a substantial improvement of the article.


  \bibliographystyle{elsarticle-harv} 
  \bibliography{references}





\end{document}